\documentclass[nohyperref]{article}

\usepackage{microtype}
\usepackage{graphicx}
\usepackage{subfigure}
\usepackage{booktabs} 

\usepackage{hyperref}

\usepackage[accepted]{icml2022}

\usepackage{amsmath}
\usepackage{amssymb}
\usepackage{mathtools}
\usepackage{amsthm}

\usepackage{bm}
\usepackage{wrapfig}
\usepackage{makecell}
\usepackage{enumitem}
\usepackage{subfigure}

\usepackage[capitalize,noabbrev]{cleveref}

\theoremstyle{plain}
\newtheorem{theorem}{Theorem}[section]
\newtheorem{proposition}[theorem]{Proposition}

\theoremstyle{definition}
\newtheorem{definition}[theorem]{Definition}

\theoremstyle{remark}

\usepackage[textsize=tiny]{todonotes}

\icmltitlerunning{Self-Organized Polynomial-Time Coordination Graphs}

\begin{document}

\twocolumn[
\icmltitle{Self-Organized Polynomial-Time Coordination Graphs}



\icmlsetsymbol{equal}{*}

\begin{icmlauthorlist}
\icmlauthor{Qianlan Yang}{equal,iiis}
\icmlauthor{Weijun Dong}{equal,iiis}
\icmlauthor{Zhizhou Ren}{equal,uiuc}
\icmlauthor{Jianhao Wang}{iiis}
\icmlauthor{Tonghan Wang}{harvard}
\icmlauthor{Chongjie Zhang}{iiis}
\end{icmlauthorlist}

\icmlaffiliation{iiis}{Institute for Interdisciplinary Information Sciences (IIIS), Tsinghua University}
\icmlaffiliation{uiuc}{Department of Computer Science, University of Illinois at Urbana-Champaign}
\icmlaffiliation{harvard}{Harvard University}

\icmlcorrespondingauthor{Chongjie Zhang}{chongjie@tsinghua.edu.cn}

\icmlkeywords{Machine Learning, ICML}

\vskip 0.3in
]



\printAffiliationsAndNotice{\icmlEqualContribution} 

\begin{abstract}
	
    Coordination graph is a promising approach to model agent collaboration in multi-agent reinforcement learning. It conducts a graph-based value factorization and induces explicit coordination among agents to complete complicated tasks. However, one critical challenge in this paradigm is the complexity of greedy action selection with respect to the factorized values. It refers to the decentralized constraint optimization problem (DCOP), which and whose constant-ratio approximation are NP-hard problems. To bypass this systematic hardness, this paper proposes a novel method, named Self-Organized Polynomial-time Coordination Graphs (SOP-CG), which uses structured graph classes to guarantee the accuracy and the computational efficiency of collaborated action selection. SOP-CG employs dynamic graph topology to ensure sufficient value function expressiveness. The graph selection is unified into an end-to-end learning paradigm. In experiments, we show that our approach learns succinct and well-adapted graph topologies, induces effective coordination, and improves performance across a variety of cooperative multi-agent tasks.

\end{abstract}

\section{Introduction}

Cooperative multi-agent reinforcement learning (MARL) is a promising approach to a variety of real-world applications, such as sensor networks \citep{zhang2011coordinated, ye2015multi}, traffic light control \citep{van2016coordinated}, and multi-robot formation \citep{alonso2017multi}. One of the long-lasting challenges of cooperative MARL is how to organize coordination for a large multi-agent system. In recent years, the popular MARL algorithms achieve scalability by deploying fully decentralized control policies, i.e., the behavior of each agent only depends on its local observation history. Due to the lack of explicit coordination, these methods may suffer from a game-theoretic pathology called relative overgeneralization \citep{panait2006biasing, wei2016lenient, bohmer2020deep}: if the values of coordinated and uncoordinated actions are not distinguished during exploration, the learning agents can jointly converge to suboptimal behaviors. Coordination graph \citep{guestrin2001multiagent} provides a promising class of algorithms to address this issue. It conducts a graph-based value factorization to explicitly represent coordination relations among agents. Each vertex in the graph corresponds to an agent, and each (hyper-) edge defines a local utility function over the joint action space of the connected agents. With such a higher-order value factorization, agents communicate to jointly optimize their actions, which helps to prevent relative overgeneralization. Despite this advantage, coordination graph would require more complex greedy action selection and extra communication cost than using fully decentralized policies. This paper mainly focuses on the challenges in greedy action selection of coordination graph.

One fundamental problem for coordination graphs is the trade-off between the representational capacity of value functions and the computational complexity of policy execution, where the latter would hinder the scalability of these algorithms. To obtain value functions with high expressiveness, an advanced approach, deep coordination graphs \citep[DCG;][]{bohmer2020deep}, considers a static complete graph connecting all pairs of agents. This graph structure has high representational capacity in terms of function expressiveness but raises a challenge for computation in the execution phase. The greedy action selection over a coordination graph can be formalized to a decentralized constraint optimization problem (DCOP), finding the maximum-value joint actions \citep{guestrin2001multiagent,  zhang2013coordinating, bohmer2020deep}. Note that the DCOP and its approximation are NP-hard problems, especially for the complete graphs used in DCG \citep{dagum1993approximating, park2004complexity}.
From the perspective of theoretical computer science, any scalable heuristic algorithm (e.g., max-sum algorithm \citep{pearl1988probabilistic}) for DCOP may have an uncontrollable gap with the optimal solution.

\begin{figure}
    \centering
    \includegraphics[width=0.77\linewidth]{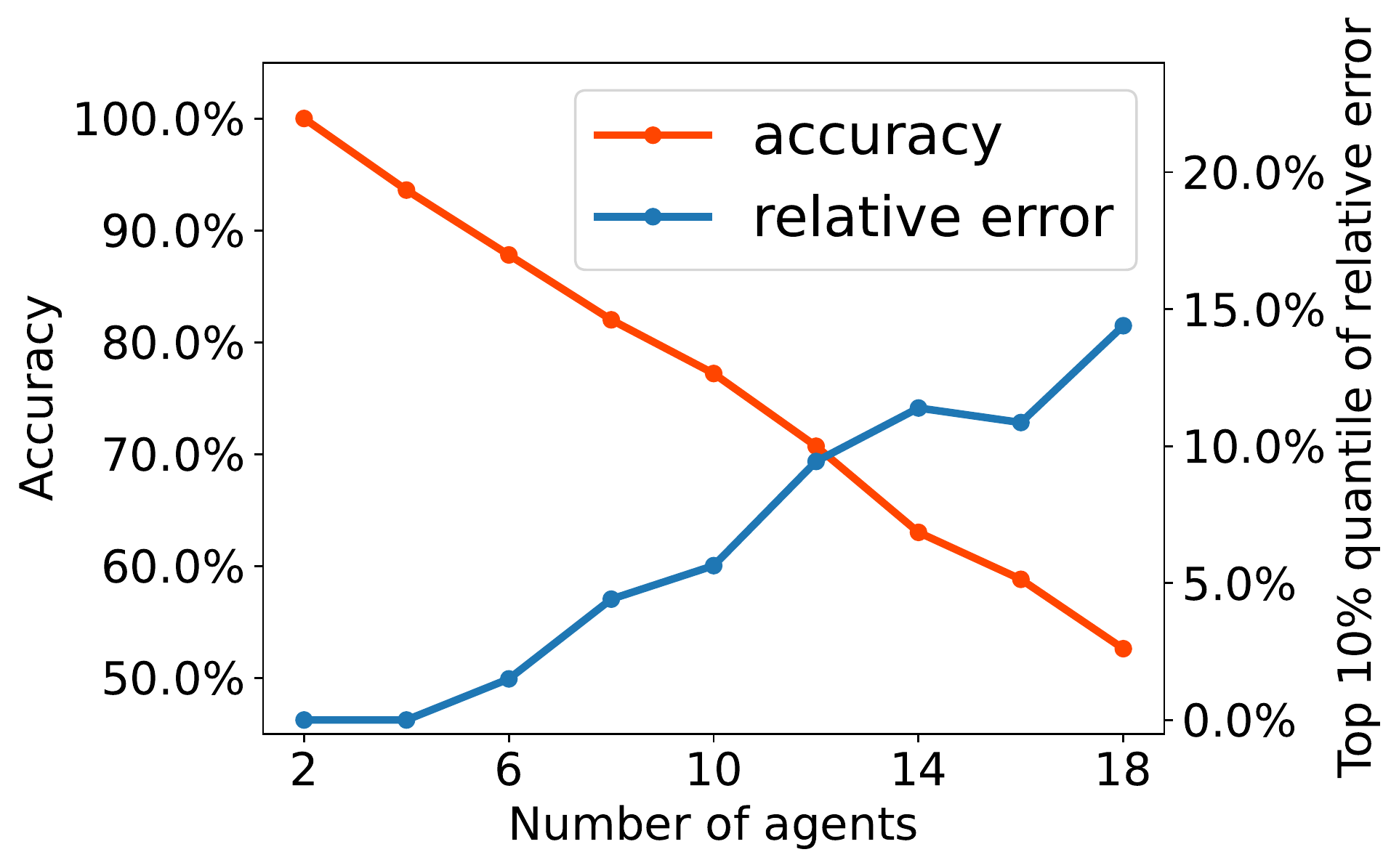}
    \vspace{-0.2in}
    \caption{The accuracy and the relative joint Q-error of the max-sum algorithm w.r.t. the number of agents.}
    \label{figure:accuracy_of_max-sum}
    \vspace{-0.2in}
\end{figure}

To visualize this issue, we generate a suite of complete-graph DCOPs with random edge-values. Fig.~\ref{figure:accuracy_of_max-sum} presents the accuracy and the relative joint Q-error of the max-sum algorithm in computing optimal joint actions, which is the default implementation of MARL methods based on coordination graphs \citep{stranders2009decentralised, zhang2013coordinating, bohmer2020deep}. As the number of agents increases, the accuracy of greedy action selection consistently decreases, and the relative joint Q-error between the selected and optimal joint actions increases accordingly. More detailed experiment setting are deferred to to Appendix \ref{appendix:dcop_test}. It further refers to a dilemma in coordination graphs, i.e., expressive graph structures lead to computationally intractable DCOPs but concise graph structures are in lack of function expressiveness. An open question in MARL raises:
\begin{center}
    \textbf{How to incorporate accurate greedy action selection with sufficient function representational capacity for coordination graph algorithms?}
\end{center}

To tackle this problem, we propose a novel graph-based MARL method for multi-agent coordination, named \textit{Self-Organized Polynomial-time Coordination Graphs} (SOP-CG), that (1) utilizes structured graph topologies with polynomial-time guarantees for DCOPs and (2) extends their representation expressiveness through a dynamic graph organization mechanism. These two characteristics open up a new family of MARL algorithms. More specifically, we first construct graph classes guaranteeing polynomial-time DCOPs and then conduct a state-dependent graph selection mechanism. Such a dynamic factorization structure is achieved by communications among agents and can be self-organized through end-to-end learning.

In experiments, we evaluate SOP-CG on the MACO benchmark \citep{wang2021context}, particle environment \citep{lowe2017multi} and StarCraft II \citep{samvelyan19smac}. SOP-CG demonstrates superior performance on tasks characterizing relative overgeneralization pathology, highlighting the effectiveness of polynomial-time coordination graphs. By extensive ablation studies, we verify that the dynamic graph selection mechanism is critical for our performance. Furthermore, we show that SOP-CG can learn interpretable and context-dependent coordination graphs, which induces effective and dynamic coordination among agents.
\section{Related Work}
Multi-agent reinforcement learning \citep{oroojlooyjadid2019review} is challenged by the size of joint action space, which grows exponentially with the number of agents. Independent Q-learning  \citep{tan1993multi,foerster2017stabilising} models agents as independent learners, which makes the environment non-stationary in the perspective of each agent. An alternative paradigm called centralized training and decentralized execution \citep[CTDE;][]{kraemer2016multi} is widely used in both policy-based and value-based methods. Policy-based multi-agent reinforcement learning methods use a centralized critic to compute gradient for the local actors \citep{lowe2017multi,foerster2018counterfactual,wen2019probabilistic,wang2020off}. Value-based methods usually decompose the joint value function into individual value functions under the IGM (individual-global-max) principle, which guarantees the consistency between local action selection and joint action optimization \citep{sunehag2018value, rashid2020monotonic, son2019qtran, wang2020QPLEX, rashid2020weighted}. Other work also studies this problem from the perspective of agent roles and individuality \citep{wang2020multi, wang2020rode, jiang2021emergence} or communication learning \citep{singh2018learning, das2019tarmac, wang2019learning}. Compared to these methods, our work is built upon graph-based value decomposition, which explicitly models the interaction among agents.

Coordination graphs are classical technique for planning in multi-agent systems \citep{guestrin2001multiagent,guestrin2002context}. They are combined with multi-agent deep reinforcement learning by recent work \citep{castellini2019representational,bohmer2020deep, li2020deep, wang2021context}. Joint action selection on coordination graphs can be modeled as a decentralized constraint optimization problem (DCOP), and previous methods compute approximate solutions by message passing among agents \citep{pearl1988probabilistic}. As the work which is most closed to our method, \citet{zhang2013coordinating} presents an algorithm based on coordination graph searching. They define a measurement to quantify the potential loss of the lack of coordination between agents and search for a coordination structure to minimize the communication cost within restricted loss of utilities. However, their induced DCOPs still remain NP-hard. In contrast, minimizing communication is not our core motivation, and we aim to use a structured graph class to maintain sufficient function expressiveness when bypassing the computational hardness of large-scale DCOPs \citep{dagum1993approximating, park2004complexity}.

\section{Background}
We consider cooperative multi-agent tasks that can be modelled as a Dec-POMDP~\citep{oliehoek2016concise} defined as $\mathcal{M} = \langle D, S, \{A^i\}_{i=1}^n, T, \{O^i\}_{i=1}^n,\{\sigma^i\}_{i=1}^n,R,h,b_0, \gamma \rangle$, where $D = \{1,\ldots,n\}$ is the set of $n$ agents, $S$ is a set of states, $h$ is the horizon of the environment, $\gamma\in[0,1)$ is the discount factor, and $b_0 \in \Delta(S)$ denotes the initial state distribution. At each stage $t$, each agent $i$ takes an action $a_{i} \in A^i$ and forms the joint action $\bm{a}=(a_{1}\ldots, a_{n})$, which leads to a next state $s'$ according to the transition function $T(s'|s,\bm{a})$ and an immediate reward $R(s,\bm{a})$ shared by all agents. Each agent $i$ observes the state only partially by drawing observations $o_{i} \in O^i$, according to $\sigma_i$. The joint history of agent $i$'s observations $o_{i,t}$ and actions $a_{i,t}$ is denoted as $\tau_{i,t} = (o_{i,0},a_{i,0},\ldots,o_{i,t-1},a_{i,t-1},o_{i,t}) \in \left(O^i \times A^i\right)^t  \times O^i$.

\paragraph{Deep Q-Learning.} Q-learning is a well-known algorithm to find the optimal joint action-value function $Q^*(s,\bm{a}) = r(s, \bm{a}) + \gamma \mathbb{E}_{s'}[ \max_{\bm{a}'} Q^*(s', \bm{a}')]$. Deep Q-learning approximates the action-value function with a deep neural network with parameters $\bm{\theta}$. In Multi-agent Q-learning algorithms~\citep{sunehag2018value,rashid2020monotonic,son2019qtran,wang2020QPLEX}, a replay memory $D$ is used to store the transition tuple $(\bm{\tau}, \bm{a}, r, \bm{\tau}')$, where $r$ is the immediate reward when taking action $\bm{a}$ at joint action-observation history $\bm{\tau}$ with a transition to $\bm{\tau}'$. $Q(\bm{\tau}, \bm{a};\bm{\theta})$ is used in place of $Q(s, \bm{a};\bm{\theta})$ because of partial observability. Hence, parameters $\bm{\theta}$ are learnt by minimizing the following expect TD error:
\begin{equation}
\mathcal{L}(\bm{\theta}) = \mathbb{E}_{(\bm{\tau},\bm{a},r,\bm{\tau}') \in D} \left[  \left( r + \gamma V\left(\bm{\tau}'; \bm{\theta}^- \right)   - Q\left( \bm{\tau}, \bm{a};\bm{\theta}\right) \right)^2 \right]
\end{equation}
where $V(\bm{\tau}';\bm{\theta}^-)=\max_{\bm{a}'}Q(\bm{\tau}',\bm{a}';\bm{\theta}^-)$ is the one-step expected future return of the TD target and $\bm{\theta}^-$ are the parameters of the target network, which will be periodically updated with $\bm{\theta}$.

\paragraph{Coordination graphs.} An undirected coordination graph \citep[CG;][]{guestrin2001multiagent} $\mathcal{G} = \langle \mathcal{V}, \mathcal{E} \rangle$ contains  vertex $v_i \in \mathcal{V}$ for each agent $1 \le i \le n$ and a set of (hyper-)edges in $\mathcal{E} \subseteq 2^{\mathcal{V}}$ which represents coordination dependencies among agents. Prior work considers higher order coordination where the edges depend on actions of several agents \citep{guestrin2002coordinated, kok2006collaborative, guestrin2002context}. Such a coordination graph induces a factorization of global Q function:
\begin{equation}
\begin{aligned}
    &Q_{tot}(s,\bm{a}) =  \sum_{v_i \in \mathcal{V}} f_i(s,a_i)\\ &+\sum_{\bm{e} =\{e_1, \ldots, e_{|\bm{e}|}\} \in \mathcal{E}} f_{\bm{e}}(s,a^{e_1}, \ldots, a^{e_{|\bm{e}|}}),
\end{aligned}
\end{equation}


where $f_i$ represents the individual utility of agent $i$ and $f_{\bm{e}}$ specifies the payoff contribution for the actions of the agents connected by the (hyper-)edge $\bm{e}$, so that the global optimal solution can be found through maximizing this joint value. The special case that $\mathcal{E}$ is an empty set yields VDN \citep{sunehag2018value}, but each additional edge enables the value representation of the joint actions of a pair of agents and can thus help to avoid relative-overgeneralization~\citep{bohmer2020deep}. In many coordination graph learning works ~\citep{zhang2013coordinating,bohmer2020deep, wang2021context}, the hyper-edges are simplified into pairwise edges. The graph is usually considered to be specified before training.  \citet{guestrin2002context} and \citet{zhang2013coordinating} suggest that the graph could also depend on states, which means each state can have its own unique CG.

\section{Self-Organized Polynomial-Time Coordination Graphs}

\begin{figure*}[t]
    \centering
    \includegraphics[width=\linewidth]{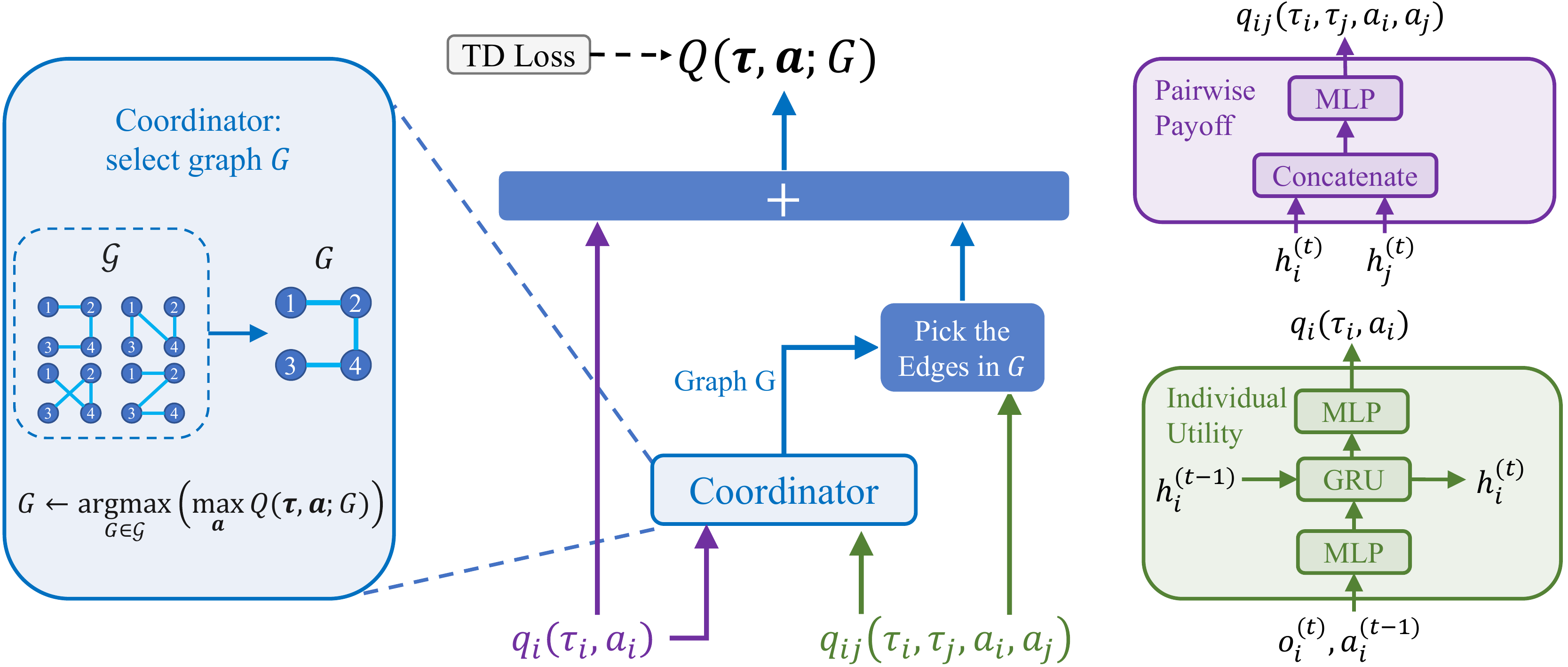}
    \vspace{-0.1in}
    \caption{An overview of SOP-CG’s framework. \textbf{Left}: An imaginary coordinator agent that selects graph $G$ from the predefined graph class $\mathcal{G}$. \textbf{Middle}: The overall working flow. \textbf{Right}: Individual utility network structure~(bottom) and pairwise payoff network structure~(top).}
    \label{figure:method}
\end{figure*}

A common method for multi-agent reinforcement learning is to decompose the joint value function into a linear combination of local value functions, each of which conditions on actions of a subset of agents \citep{guestrin2001multiagent, bohmer2020deep}. With this paradigm, computing joint actions with maximum value can be modeled as a distributed constraint optimization problem (DCOP). General DCOP 
and any constant-ratio approximation have been proved to be NP-hard \citep{dagum1993approximating, park2004complexity}. Previous work adopts a variety of heuristic algorithms (e.g., message passing methods) for action selection \citep{kok2006collaborative, wang2021context}. The imperfection of heuristics may lead to two side effects: (a) collecting bad samples and (b) errors in constructing one-step TD target, which hurt the learning performance.

To address these issues, we investigate the \emph{polynomial-time coordination graph class}, in which the induced DCOPs can be solved in polynomial time. We present a novel algorithm called \emph{Self-Organized Polynomial-time Coordination Graphs} (SOP-CG), that utilizes a class of polynomial-time coordination graphs to construct a dynamic and state-dependent topology. In Section~\ref{method:value_decomposition}, we introduce our value factorization upon specific coordination graphs, which can hold the accurate greedy action selection without significantly sacrificing the representational capacity by using a bottom-up design. To enable efficient topology self-organization, in Section~\ref{method:framework}, we unify the graph selection into the Q-learning framework by formulating the graphs as actions of an imaginary coordinator agent, and depict the whole framework of our algorithm. In Section~\ref{section:graph_classes}, we further propose two polynomial-time graph class instances and discuss their theoretical advantages in SOP-CG. Fig.~\ref{figure:method} illustrates the overall framework of our approach.

\subsection{Value Factorization on Polynomial-Time Coordination Graphs}
\label{method:value_decomposition}

Our approach first models coordination relations upon the graph-based value factorization specified by deep coordination graphs \citep[DCG;][]{bohmer2020deep}, in which the joint value function of the multi-agent system is factorized to the summation of \textit{individual utility functions} $q_i$ and \textit{pairwise payoff functions} $q_{ij}$ as follows:
\begin{equation}
\begin{aligned}
&Q(\boldsymbol{\tau}^{(t)},\boldsymbol{a};G) = \sum_{i\in[n]} q_i(\tau_i^{(t)},a_i) \\
&+ \sum_{(i,j)\in G}q_{ij}(\tau_i^{(t)},\tau_j^{(t)},a_i,a_j),
\label{equation:qtot}
\end{aligned}
\end{equation}
where the coordination graph $G$ is represented by a set of undirected edges. With this second-order value decomposition, the hardness of greedy action selection is highly related to the graph topology. To provide a clear perspective, we discuss the theoretical complexity of the DCOPs induced by different topology classes.

\begin{proposition}
    Let $n$ be the number of agents and $A=|\bigcup_{i=1}^n A^i|$. (i) Approximating the induced DCOPs of fully connected coordination graphs within a factor of $O\left(\frac{\log A}{\sqrt A}\right)$ is NP-hard. (ii) There exists an algorithm that can compute the accurate solution of any DCOP induced by an undirected acyclic graph in $\text{Poly}(n, A)$ running time.
    \label{proposition:dcop_complexity}
\end{proposition}

Detailed proof can be found in Appendix~\ref{appendix:proof}. Notebly, DCG adopts the fully connected coordination graph as the default implementation. Due to the theoretical hardness of its induced DCOP, using any polynomial-time approximation algorithm may result in unfavorable quality of the selected actions. To avoid this problem, we propose to deploy \textit{Polynomial-Time Coordination Graph Classes}. Now we formulate the definition as follows:


\begin{definition}[Polynomial-Time Coordination Graph Class]
    We say a graph class $\mathcal{G}$ is a \textit{Polynomial-Time Coordination Graph Class} if there exists an algorithm that can solve any induced DCOP of any coordination graph $G \in \mathcal{G}$ in $\text{Poly}(n, A)$ running time.
\end{definition}


As stated in Proposition~\ref{proposition:dcop_complexity}, the set of undirected acyclic graphs $\mathcal{G}_{\text{uac}}$ is a polynomial-time coordination graph class.
However, an undirected acyclic graph can contain at most $n-1$ edges in an environment with $n$ agents, which suffers from the lack of function expressiveness.


To alleviate this problem, our approach allows the graph topology to dynamically evolve through the transitions of environment status. Given different environmental states, the joint values can be factorized with different coordination graphs chosen from a predefined graph class $\mathcal{G}\subseteq \mathcal{G}_{\text{uac}}$. This design is based on the assumption that, although a long-horizon task cannot be characterized by a static sparse coordination graph, the coordination relations at each single-time step are sparse and manageable. The employment of this graph class fills the lack of representational capacity as well as maintains the accuracy of greedy action selections.



\subsection{Learning Self-Organized Topology with An Explicit Coordinator}
\label{method:framework}

Now we present a novel framework to render the self-organized coordination graph. We introduce an imaginary coordinator agent whose action space refers to the selection of graph topologies, aiming to select a proper graph for minimizing the suboptimality of performance within restricted coordination. When using the graph-based value factorization stated in Eq.~\eqref{equation:qtot}, the graph topology $G$ can be regarded as an input of joint value function $Q({\bm \tau}^{(t)},{\bm a};G)$. The objective of the coordinator agent is to maximize the joint value function over the specific graph class, which is a dual problem for finding a graph to minimize the suboptimality of performance within the different restricted coordination structures \citep{zhang2013coordinating}. Under this interpretation, we can integrate the selection of graph topologies into the trial-and-error loop of reinforcement learning. We handle the imaginary coordinator as a usual agent in the multi-agent Q-learning framework and rewrite the joint action as ${\bm a}_{cg}=(a_1,\cdots,a_n,G)$.

\paragraph{Execution.} Formally, at time step $t$, greedy action selection indicates the following joint action:
\begin{equation}
{\bm a}_{cg}^{(t)} \gets \mathop{\arg\max}_{(a_1,\cdots,a_n,G)} Q(\boldsymbol{\tau}^{(t)},a_1,\cdots,a_n;G).
\label{equation:joint_action}
\end{equation}

Hence the action of coordinator agent $G^{(t)}$ is naturally the corresponding component in Eq.~\eqref{equation:joint_action}:
\begin{equation}
G^{(t)} \gets \mathop{\arg\max}_{G\in\mathcal{G}} \left( \max_{\boldsymbol{a}} Q(\boldsymbol{\tau}^{(t)},\boldsymbol{a};G) \right).
\label{equation:centralized_coordinator}
\end{equation}
After determining the graph topology $G^{(t)}$, the agents can choose their individual actions to jointly maximize the value function $Q({\bm \tau}^{(t)},{\bm a};G^{(t)})$ upon the selected topology.

The imaginary agent is only introduced to better demonstrate our idea. In practice, it is not necessary to explicitly employ a coordinator. The agents can jointly select the graph topology and coordinate their actions through communications. This communication mechanism is consistent with previous coordination-graph-based methods like DCG and CASEC \citep{wang2021context}.

\paragraph{Training.} With the imaginary coordinator, we can re-formulate the Bellman optimality equation and maximize the future value over the coordinator agent's action:
\begin{equation}
Q^*(\boldsymbol{\tau},\boldsymbol{a};G)=\mathop\mathbb{E}_{\boldsymbol{\tau}'}\left[r+\gamma\max_{G'}\max_{\boldsymbol{a}'} Q^*(\boldsymbol{\tau}',\boldsymbol{a}';G')\right]
\label{equation:bellman}
\end{equation}

Since the graph selection is a part of agent action, the associated value $Q({\bm\tau},{\bm a};G)$ of graph $G$ can be updated through temporal difference learning. The network parameters $\bm{\theta}$ can be trained by minimizing the standard Q-learning TD loss:
\begin{equation}
\mathcal{L}_{cg}(\bm{\theta})=\mathop{\mathbb{E}}_{(\boldsymbol{\tau},\boldsymbol{a},G,r,\boldsymbol{\tau}')\sim \mathcal{D}}\left[\left(y_{cg}-Q(\boldsymbol{\tau},\boldsymbol{a};G;\bm{\theta})\right)^2\right]
\label{equation:td_loss}
\end{equation}
where $y_{cg}=r+\gamma\max_{(\bm{a}',G')}Q(\boldsymbol{\tau}',\boldsymbol{a}';G';\bm{\theta}^-)$ is the one-step TD target and $\bm{\theta}^{-}$ are the parameters of target network. $G^{(t)}$ is regarded as part of the transition and is stored in the replay buffer together with the realistic agent actions. We include more implementation details of temporal difference learning in Appendix~\ref{appendix:graph_relabel}.

The overall algorithm of our approach is summarized in Algorithm \ref{algorithm:framwork}. The graph selection step is the main characteristic of our method, which differs from the static graph representation used by prior work \citep{guestrin2002coordinated,kok2006collaborative,bohmer2020deep}. To support the self-organization of coordination graphs, the graph selection step raises a new computation demand, i.e., we need to find the best candidate from the graph class $\mathcal{G}$, which will be another non-trivial combinatorial optimization problem. We will show that, by designing a proper graph class $\mathcal{G}$, we can maintain both computational efficiency and function representational capacity in the next subsection.

\begin{algorithm}[tb]
	\caption{Self-Organized Polynomial-Time Coordination Graphs}
	\begin{algorithmic}
		\REQUIRE Predefined graph class $\mathcal{G}$
		\STATE Initialize replay buffer $\mathcal{D}$
		\STATE Initialize network with random weights
		\FOR {episode$=1,\cdots,M$}
		\STATE Receive initial observation $\boldsymbol{\tau}^{(0)}$
		\FOR {$t=0,\cdots,T$}
		\STATE Choose coordination graph $G^{(t)}\in \mathcal{G}$ according to Eq.~\eqref{equation:centralized_coordinator} 
		\STATE Select actions $\boldsymbol{a}^{(t)}$ w.r.t $G^{(t)}$ and an $\epsilon$-greedy exploration
		\STATE Execute actions $\boldsymbol{a}^{(t)}$ and receive reward $r^{(t)}$ and observation $\boldsymbol{o}^{(t+1)}$
		\STATE Store transition in replay buffer $\mathcal{D}$
		\ENDFOR
		\STATE Sample random minibatch of transitions from $\mathcal{D}$
		\STATE Perform a gradient descent step on loss defined in Eq.~\eqref{equation:td_loss}
		\ENDFOR
	\end{algorithmic}
	\label{algorithm:framwork}
\end{algorithm}


\begin{figure*}[htb]
    \centering
    \begin{minipage}{0.32\textwidth}
        \centering
        \includegraphics[height=3.2cm]{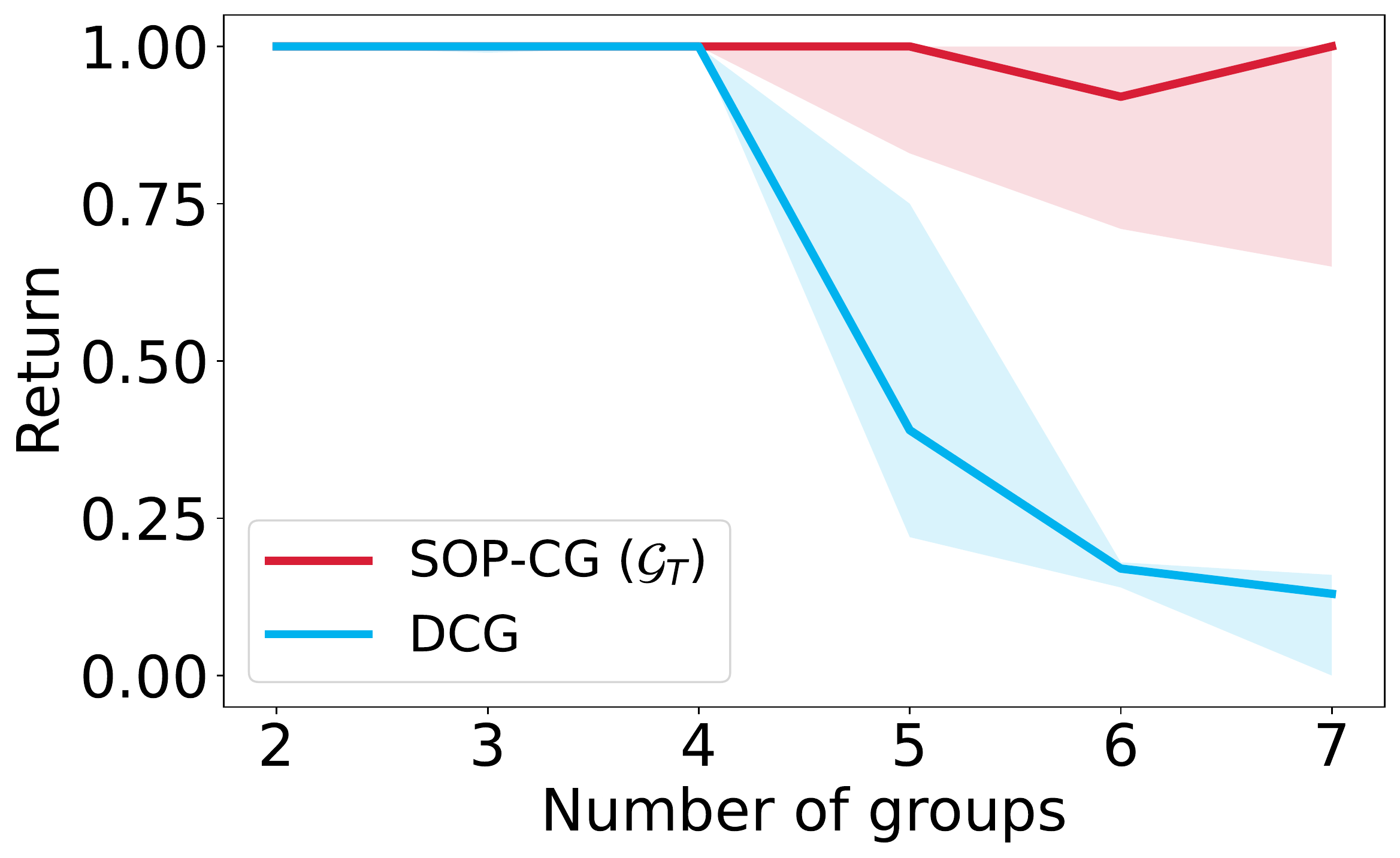}
    \end{minipage}
    \begin{minipage}{0.33\textwidth}
        \centering
        \includegraphics[height=3.2cm]{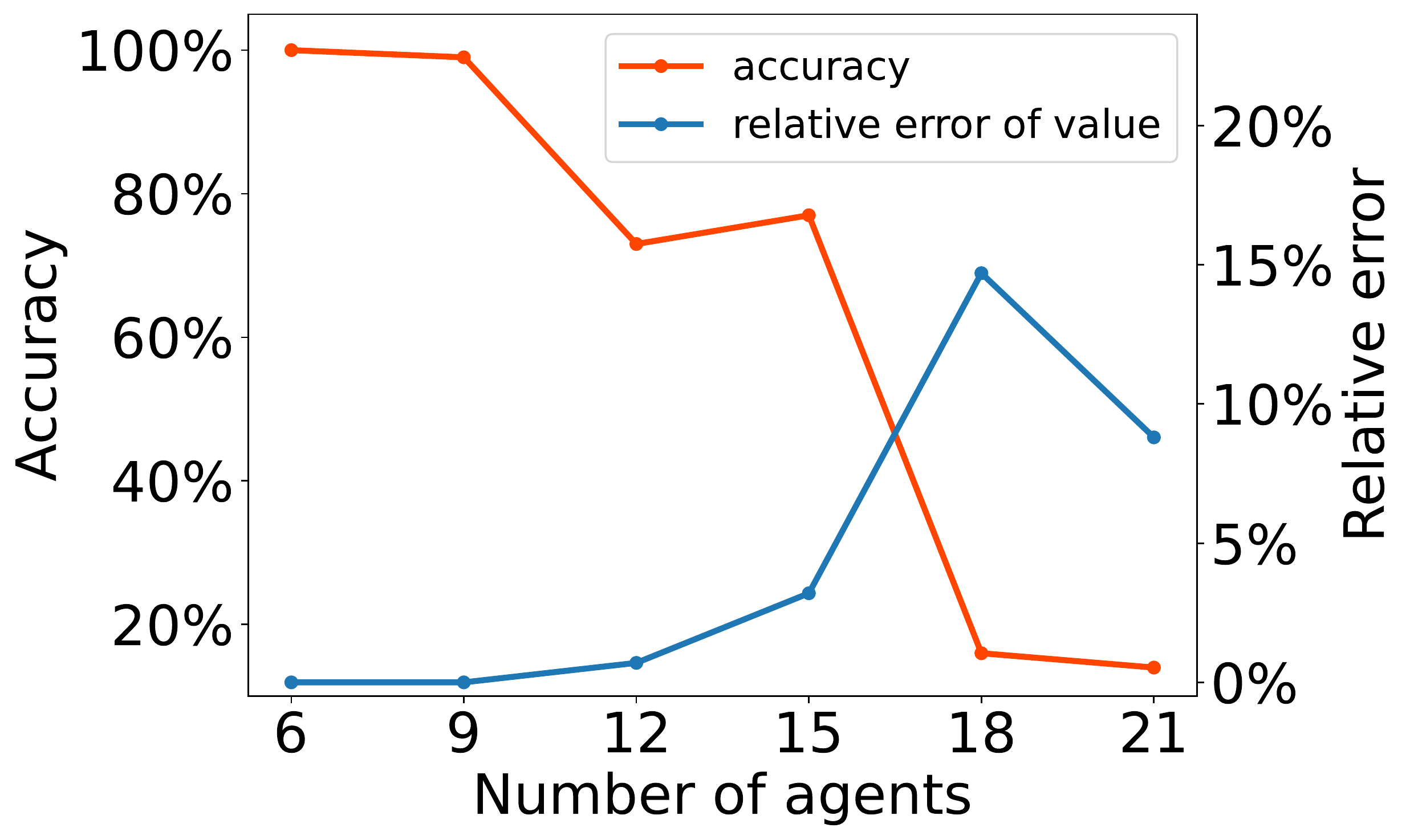}
    \end{minipage}
    \begin{minipage}{0.33\textwidth}
        \centering
        \includegraphics[height=3.2cm]{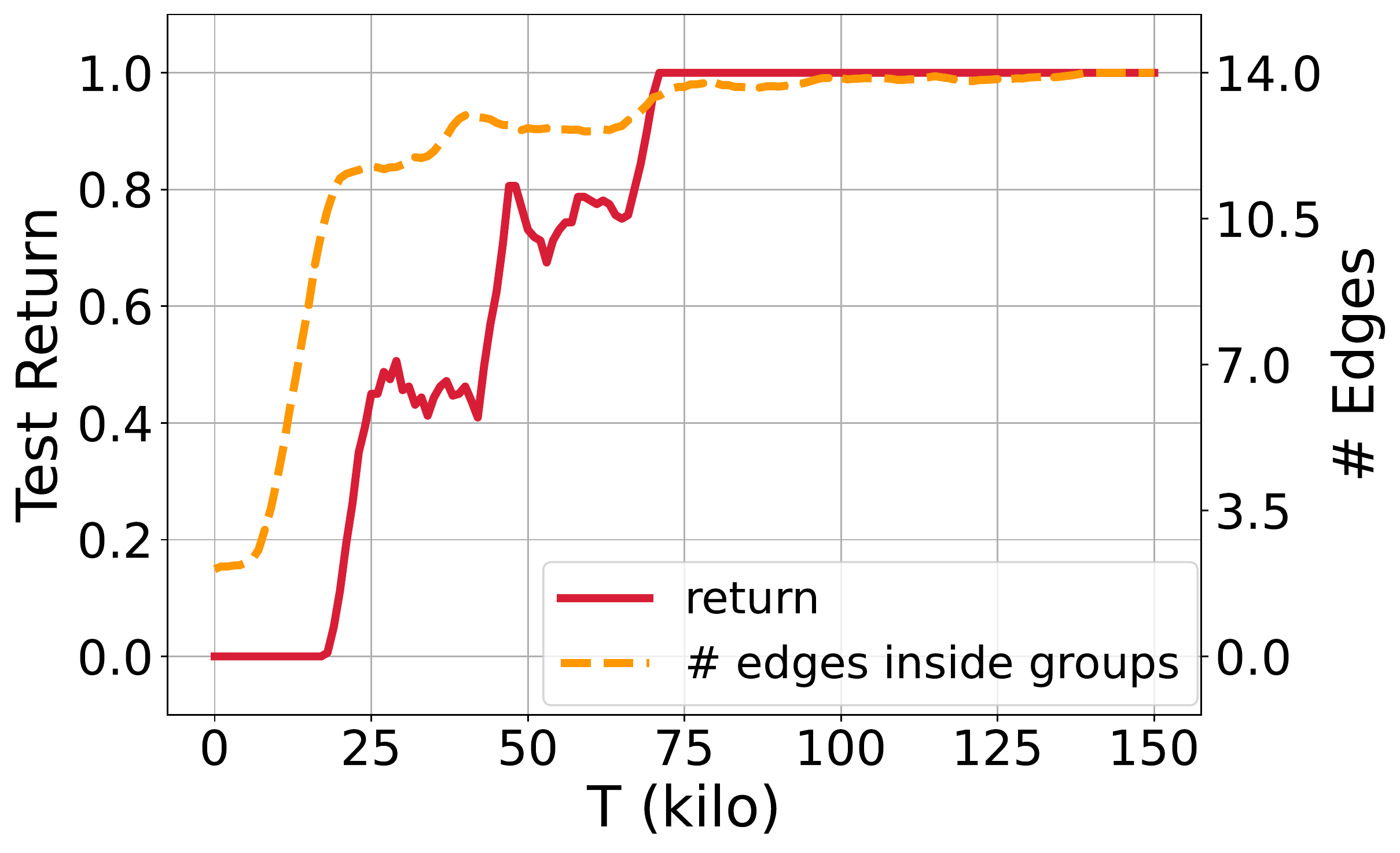}
    \end{minipage}
    \vspace{-0.05in}
    \caption{Illustrative example. \textbf{Left}: Final performance of SOP-CG ($\mathcal{G}_T$) and DCG when $k=2$ to $7$. The optimal return is 1. \textbf{Middle}: Accuracy of DCG's action selection, and the relative error between the joint values computed by the taken actions and the optimal actions. \textbf{Right}: Number of edges inside groups selected by SOP-CG ($\mathcal{G}_T$) when $k=7$. A graph in $\mathcal{G}_T$ can include at most 14 edges which connects agents in the same group.}
    \label{figure:illustrating_example}
    \vspace{-0.05in}
\end{figure*}

\subsection{Harnessing Structured Graph Classes for Efficient Coordination} \label{section:graph_classes}

To achieve the efficient graph search over the polynomial-time class instances, we investigate the following structures for the graph class $\mathcal{G}$, which are subsets of $\mathcal{G}_{\text{uac}}$:
\begin{itemize} \setlength{\itemsep}{0.9pt}
    \item \emph{Pairwise grouping $\mathcal{G}_P$}. This class is introduced by \cite{castellini2019representational}. $n$ agents are partitioned to $\lfloor n/2\rfloor$ non-overlapping pairwise groups. Only pairwise interactions are considered and two agents are connected with an edge if they lie in the same group.
    \item \emph{Tree organization $\mathcal{G}_T$}. Overlapping edges are allowed in this class. In a system with $n$ agents, the graph is organized as a tree with $n-1$ edges so that all agents form a connected component. Even if two agents are not directly connected by an edge, their actions are implicitly coordinated through the path on the tree. This provides the potential to approximate complex joint value functions.
\end{itemize}

\begin{table}[t]
    \centering
    \begin{tabular}{crcrcr}
        \toprule
        \multicolumn{2}{c}{Graph Class} & \multicolumn{2}{c}{\makecell[c]{Select\\graph $G^{(t)}$}} & \multicolumn{2}{c}{\makecell[c]{Select actions on\\a given graph}} \\
        \cmidrule(lr){1-2} \cmidrule(lr){3-4} \cmidrule(lr){5-6}
        \multicolumn{2}{c}{$\mathcal{G}_P$} & \multicolumn{2}{c}{$\surd$} & \multicolumn{2}{c}{$\surd$} \\
        \cmidrule(lr){1-2} \cmidrule(lr){3-4} \cmidrule(lr){5-6}
        \multicolumn{2}{c}{$\mathcal{G}_T$} & \multicolumn{2}{c}{$-$} & \multicolumn{2}{c}{$\surd$} \\
        \cmidrule(lr){1-2} \cmidrule(lr){3-4} \cmidrule(lr){5-6}
        \multicolumn{2}{c}{Complete graph} & \multicolumn{2}{c}{N/A} & \multicolumn{2}{c}{$-$} \\
        \bottomrule
    \end{tabular}
    \caption{Computational considerations in choosing graph class. A check mark ``$\surd$'' means that known algorithms for the problem are guaranteed to find the optimal solution in polynomial time. The mark ``$-$'' denotes that these settings use the specific heuristic methods.}
    \label{table:complexity}
\end{table}

$\mathcal{G}_T$ is actually the set of maximal undirected acyclic graphs. It is noteworthy that we do not further incorporate the smaller graphs into both classes, since their representational capacity are dominated by existing graphs in the classes. We summarize the hardness of computations upon different graph class in Table~\ref{table:complexity}. Although the representational capacity of $\mathcal{G}_P$ is relatively weaker, selecting $G^{(t)}$ from this class can be rigorously solved by polynomial-time algorithms \citep{edmonds1965paths}, which highlights the key benefit of $\mathcal{G}_P$. In comparison, selecting $G^{(t)}$ from $\mathcal{G}_T$ is non-trivial, so we design a greedy algorithm to compute approximated solutions. Note that, although the graph selection of $\mathcal{G}_T$ may be imprecise, its represented values are formally optimized by temporal difference learning with accurate action selection. Thus, as we will show in the experiment section, $\mathcal{G}_T$ can still work as a promising graph class in absence of rigorous graph selection. Please refer to Appendix~\ref{appendix:algorithm} for the implementation details of the induced combinatorial optimization for graph selection.

\paragraph{Remark.}
Regarding the integration of these graph classes and the maximization criterion introduced in Eq.~\eqref{equation:centralized_coordinator}, we have not derived strong convergence guarantees of the graph structure and the value functions. Despite this limitation, the criterion works well and can produce interpretable graph topologies that reflect the ground-truth problem structure, as we will show in Section~\ref{section:experiment}. A future direction is to explore other graph organization criteria specializing in theoretical properties, e.g., developing systematical analysis and improving the convergence rate.


\section{Experiments}
\label{section:experiment}

\begin{figure*}[t]
    \centering
    \includegraphics[width=\linewidth]{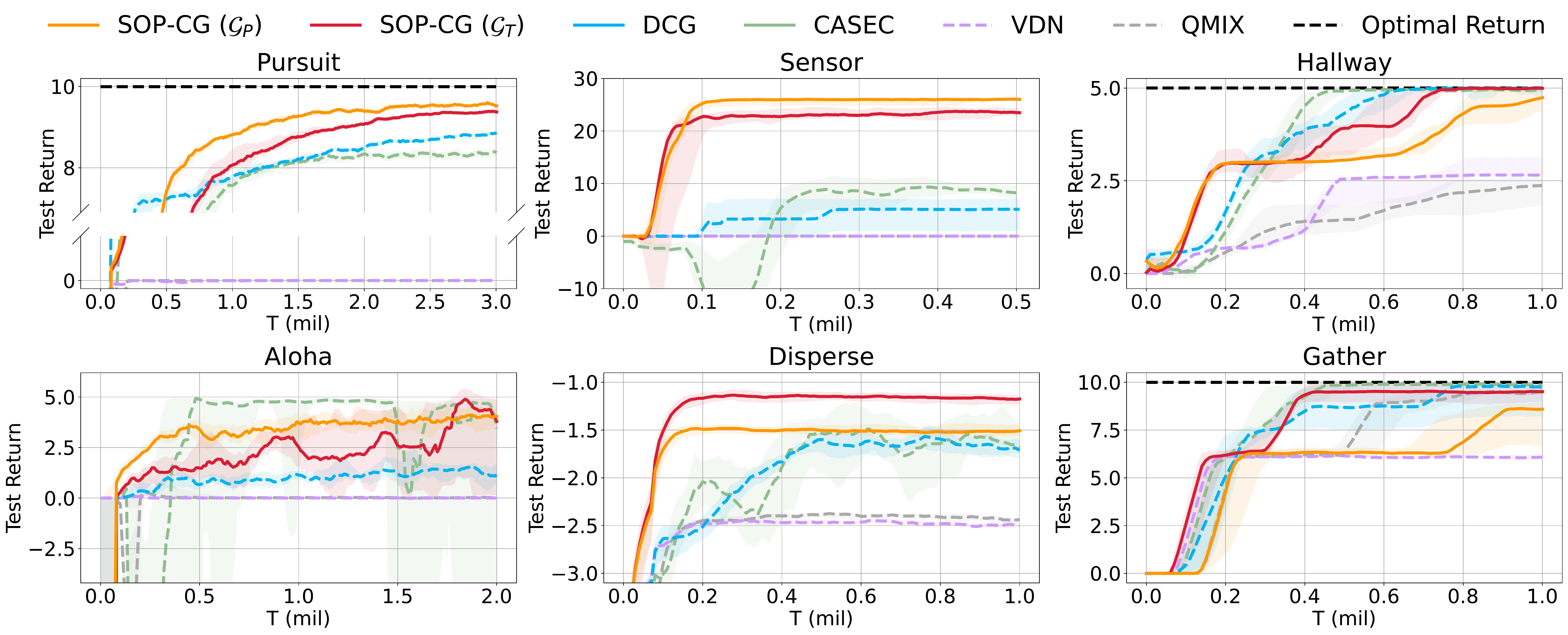}
    \vspace{-0.3in}
    \caption{Learning curves on MACO benchmark. The optimal return is demonstrated if it is available.}
    \label{figure:maco}
\end{figure*}

\begin{figure}[t]
    \centering
    \includegraphics[width=0.7\linewidth]{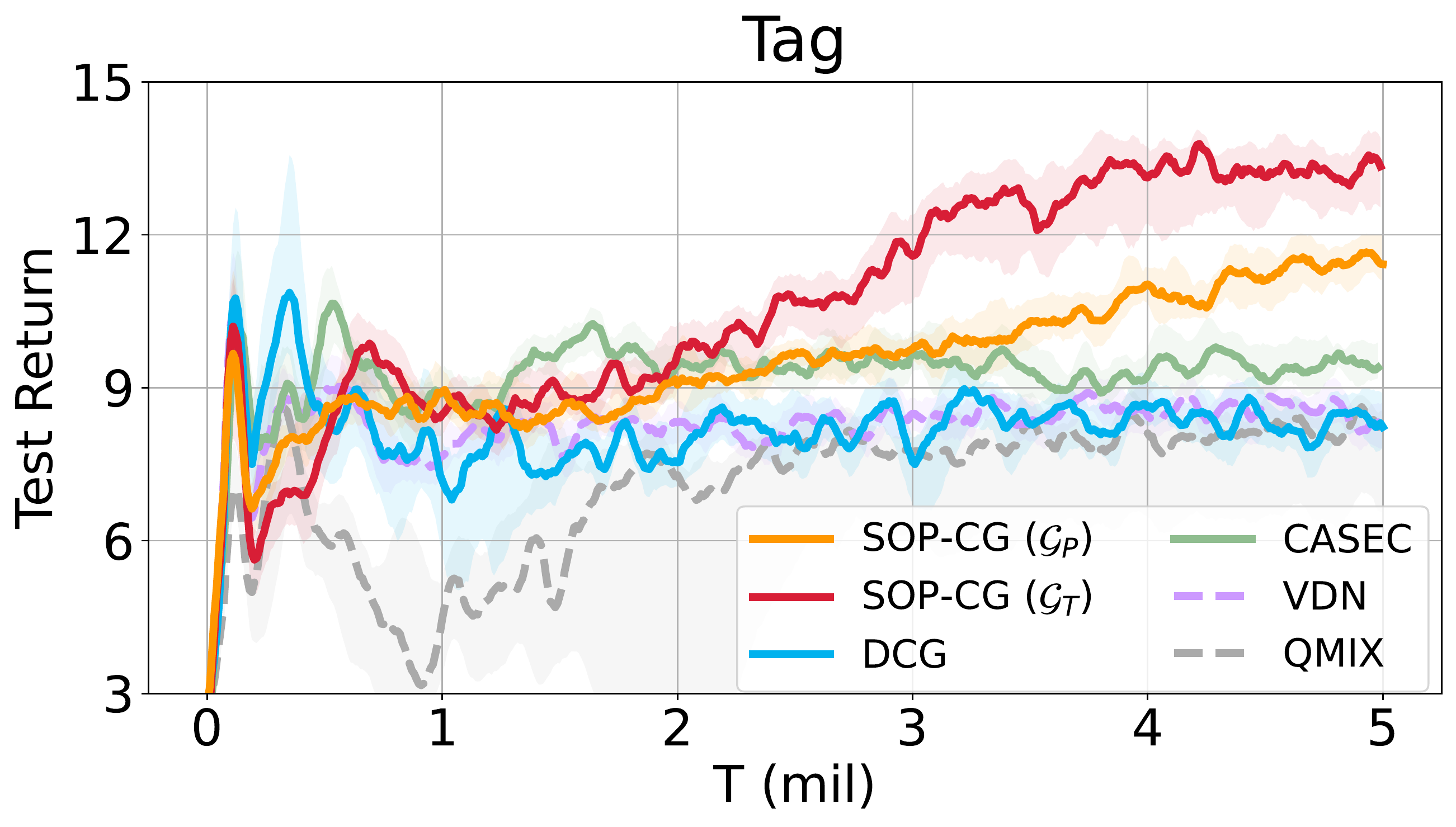}
    \vspace{-0.15in}
    \caption{Learning curves on Tag.}
    \label{figure:tag}
    \vspace{-0.15in}
\end{figure}

In this section, we conduct experiments to answer the following questions: (1) Does the accurate greedy action selection improve our performance (see Section~\ref{didactic_example} and \ref{performance_comparison})? (2) Is the dynamic graph organization mechanism necessary for our algorithm (see Section~\ref{didactic_example} and \ref{ablation})? (3) How well does SOP-CG perform on complex cooperative multi-agent tasks (see Fig.~\ref{figure:maco}, \ref{figure:tag} and \ref{figure:smac})? (4) Can SOP-CG extract interpretable dynamic coordination structures in complex scenarios (see Section~\ref{visualization})? For evaluation, all results in this section are illustrated with the median performance over 5 random seeds as well as the 25\%-75\% percentiles. An open-source implementation of our algorithm is available online\footnote{ \url{https://github.com/yanQval/SOP-CG}.}.

\subsection{Illustrative Example}
\label{didactic_example}

To illustrate the effects of the accuracy of greedy action selection in DCG and SOP-CG, we devise a simple one-step cooperative game for $3 \times k$ agents with two actions.

The $3 \times k$ agents will be randomly separated into $k$ groups with each group $3$ agents. Action A is available to all the agents, while action B is only available to a random set of agents. Choosing action B will yield a cost of 0.5. If three agents in the same group take action B, they will receive a global reward of 2.5. Therefore, agents in the same group should coordinate to decide whether to jointly take action A or action B.

We set the number of groups from $2$ to $7$ and examine the final performance of DCG and SOP-CG after $150000$ episodes training. As shown in Fig.~\ref{figure:illustrating_example}-Left, SOP-CG can keep achieving the optimal return, while DCG is more sensitive to the total number of agents and obtains low return when $k\geq 5$. For DCG, we further test the accuracy of its greedy action selection as well as the relative error between the joint values computed by the taken actions and the optimal actions. Fig.~\ref{figure:illustrating_example}-Middle demonstrates that the accuracy plummets when the number of agents grows. This problem directly limits the performance of DCG, whereas SOP-CG does not suffer from it.

Meanwhile, we empirically show that our method could learn well-adapted topologies during the training. In this game, each agent only needs to coordinate with others in the same group. To better understand the contribution of self-organized graph selection, we visualize the number of edges which connects agents in the same group in Fig.~\ref{figure:illustrating_example}-Right. It continually increases during the training, while the test performance rises along with it and finally reaches the optimal value after 14 such edges are chosen, which is the maximum number of edges inside groups that a graph in $\mathcal{G}_T$ can cover. This phenomenon proves the ability of our method to learn better policy through discovering the ground truth coordination requirements in the environment.

The full details of the illustrative example are provided in Appendix \ref{appendix:illustrative}.

\subsection{Performance Comparison}
\label{performance_comparison}

In this section, we test SOP-CG on multi-agent coordination challenge \citep[MACO;][]{wang2021context}, multi-agent particle environment \citep[MPE;][]{lowe2017multi} and StarCraft multi-agent challenge \citep[SMAC;][]{samvelyan19smac}. We compare SOP-CG with the state-of-the-art coordination graph learning methods (DCG~\citep{bohmer2020deep}, CASEC~\citep{wang2021context}) and fully decomposed value-based methods (VDN~\citep{sunehag2018value}, QMIX~\citep{rashid2020monotonic}). The implementation details and hyperparameter settings are included in Appendix~\ref{appendix:hyperparameter}.

\begin{figure}[t]
    \centering
    \includegraphics[width=1\linewidth]{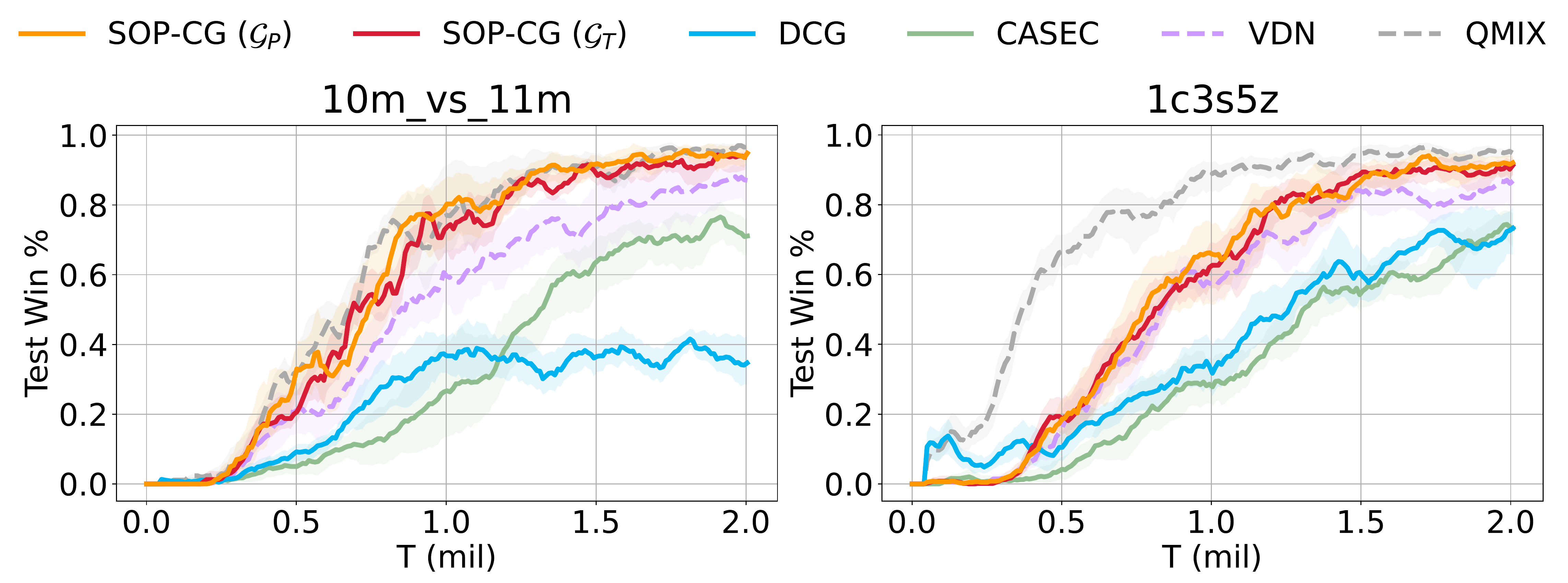}
    \vspace{-0.22in}
    \caption{Learning curves on SMAC tasks.}
    \label{figure:smac}
    \vspace{-0.1in}
\end{figure}

\begin{figure*}
    \centering
    \includegraphics[width=\linewidth]{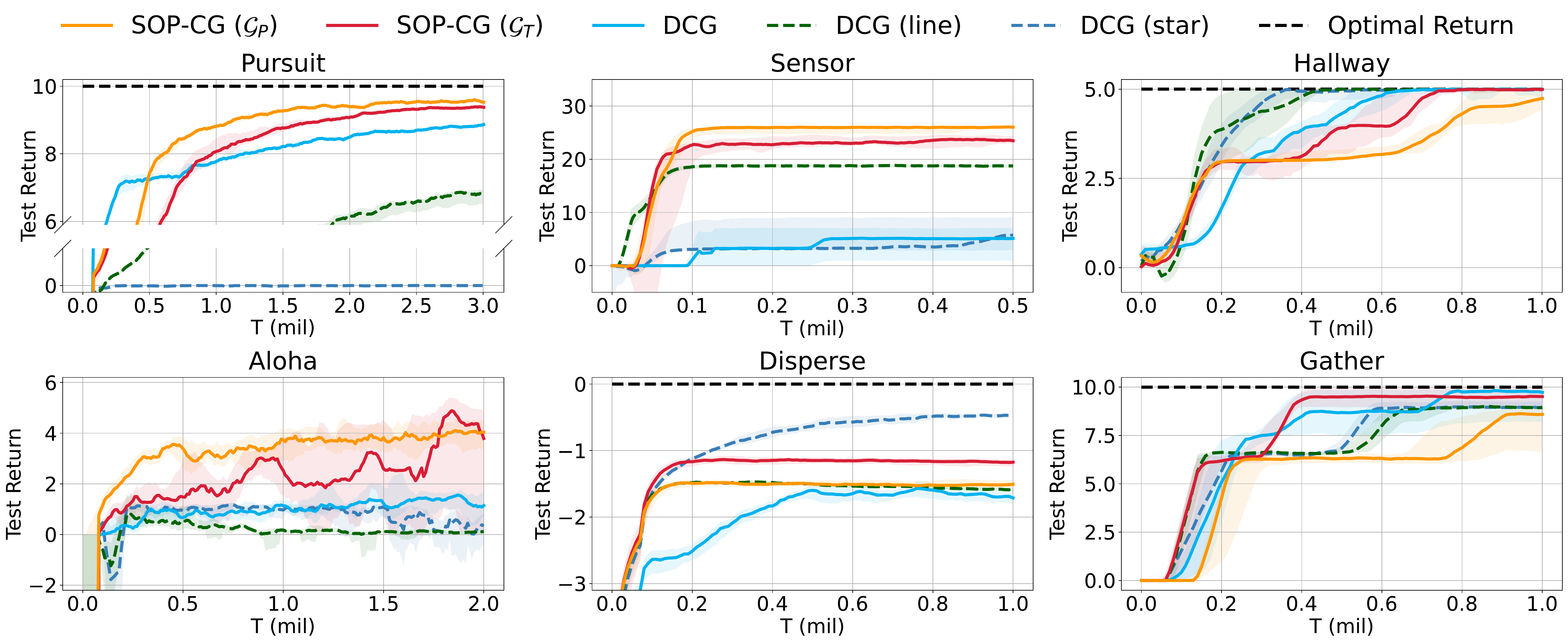}
    \vspace{-0.2in}
    \caption{Ablation study on MACO benchmark. \emph{Line} and \emph{Star} are two special instances of the polynomial-time graph class. The optimal return is demonstrated if it is available.}
    \label{figure:ablation}
\end{figure*}   

MACO \citep{wang2021context} is a benchmark consisting of 6 classic coordination problems, raising challenges of relative overgeneralization and partial observability. Results are illustrated in Fig.~\ref{figure:maco}. SOP-CG outperforms the baselines on three tasks, especially Pursuit and Sensor. Our method learns almost the optimal policy on Pursuit while baselines have over two times larger gap with the optimal return than ours. On Sensor, the return achieved by our method is over twice the best baseline's return. The major difficulties on these tasks lie in \textit{relative overgeneralization pathology} \citep{panait2006biasing, wei2016lenient, bohmer2020deep}, since the uncoordinated actions will lead to high punishments. Even if an agent takes the same action, the reward observed by it is highly related to the actions of other agents and may be drastically changeable due to their exploration, which hinders the credit assignment of fully decentralized methods. Although the explicit modeling of agent interactions helps the coordination-graph-based methods address this issue, accurate action selection is also critical for learning optimal policy. When facing growing scales (there are 20 agents on Pursuit\footnote{Pursuit is also called Predator-Prey \citep{son2019qtran, rashid2020weighted}. MACO benchmarks Pursuit with 10 agents and reward : penalty = 1 : 1. In this paper, we test a harder version with 20 agents.} and 15 agents on Sensor), DCG and CASEC have a poor accuracy on DCOP, which may lead to inaccurate greedy action selection, impeding them to better performance. In contrast, our method converges to the superior values, highlighting the effectiveness of polynomial-time coordination graphs.

Tag is a task based on the particle world \citep{lowe2017multi}. 10 agents chase 3 adversaries on the map with 3 randomly generated obstacles, and the agents receive a global reward for each collision between an agent and an adversary. Since the adversaries move faster, it requires fine-grained control and collaboration for the agents to surround the adversaries. As shown in Fig.~\ref{figure:tag}, SOP-CG ($\mathcal{G}_T$) significantly outperforms the baselines on this task. In Section~\ref{visualization}, we will visualize the coordination structures learned by our method to justify the adaptability of the dynamic graph organization mechanism.

We further test SOP-CG on the SMAC benchmark to demonstrate its scalability in large action-observation spaces. Notably, the number of output heads of pairwise payoff functions grows quadratically with the number of actions. In this environment, DCG and CASEC adopt different techniques (e.g., low-rank approximation or action representation learning) to improve the learning efficiency of the payoff functions, which can also be applied to our algorithm. For fair comparison, we equip SOP-CG and the coordination-graph-based baselines (i.e., DCG and CASEC) with action representation learning. Fig.~\ref{figure:smac} illustrates the results on the tasks 10m\_vs\_11m and 1c3s5z. SOP-CG outperforms both DCG and CASEC by a large margin.

\begin{figure*}[t]
    \centering
    \vspace{0.07in}
    \includegraphics[width=\linewidth]{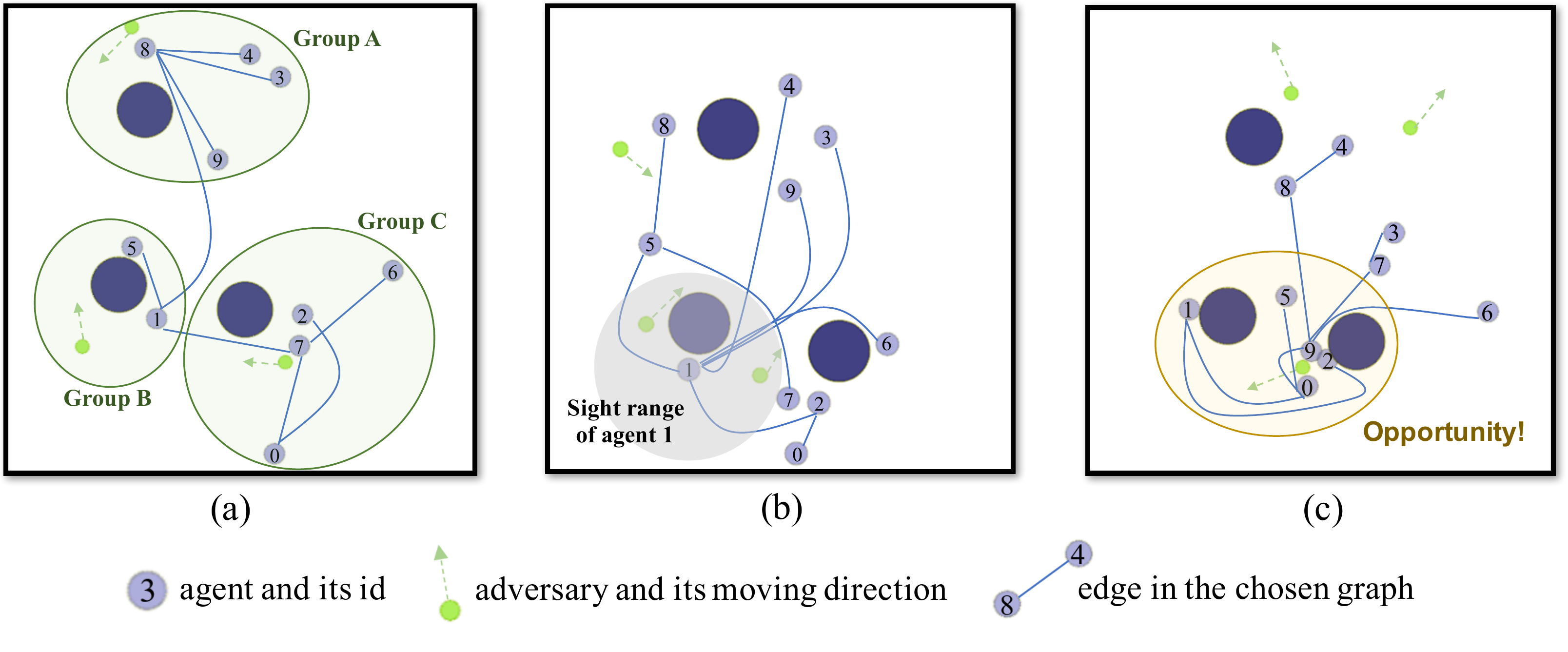}
    \vspace{-0.3in}
    \caption{Various coordination graphs learned by our algorithm on Tag: (a) Self-organized grouping at initialization; (b) Connecting to agent with rich observation for better information sharing; (c) Concentrated collaboration structure around an enclosed adversary.}
    \label{figure:tag_illustration}
\end{figure*}

\subsection{Ablation Study}
\label{ablation}

In this section, we carry out ablation studies to justify the contribution of the self-organized dynamic topology. We test alternative implementations of DCG on two static graphs \emph{Line} (agent $i$ is connected with agent $i-1$ and agent $i+1$) and \emph{Star} (agent $1$ is linked with others). These graphs are two special instances of $\mathcal{G}_{uac}$, implying polynomial-time solvable DCOPs. Results are shown in Fig.~\ref{figure:ablation}. Although the function expressiveness is restricted, it is observed that DCG (line) and DCG (star) can perform much better than DCG in some cases (i.e., DCG (line) on Sensor and DCG (star) on Disperse). We hypothesize that this is because these topologies induce efficient cooperation strategy or directly match the ground truth coordination demands on such domains, and the accurate greedy action selection ensures them to achieve superior performance. For example, agents in Sensor are arranged as a $3\times 5$ matrix in order of their ids. As adjacent agents need to coordinate to scan a nearby target, \emph{Line} covers the major coordination requirements and renders a satisfactory policy. However, these methods have poor performance on other tasks, which may not be characterized by such static coordination graphs. In contrast, SOP-CG achieves good performance across all tasks in MACO by virtue of self-adaptive topologies.

\subsection{Visualization of Self-Organized Coordination}
\label{visualization}

A major strength of our method is that SOP-CG can organize different coordination graphs to adapt to different situations. Such an adaptive organization mechanism can improve the efficiency of multi-agent collaboration. To illustrate the coordination relations organized by SOP-CG, we visualize the learned graph structures on Tag. In Fig.~\ref{figure:tag_illustration}, we take three snapshots chronologically from one episode. This episode is collected by SOP-CG ($\mathcal{G}_T$) using the tree-based graph class. It shows that SOP-CG can produce interpretable graph structures that characterize the ground-truth coordination dependencies. The learned coordination behaviors are interpreted as follows:
\begin{enumerate}
    \item Fig.~\ref{figure:tag_illustration}-a presents the initial status of the game, where agents and adversaries are generated randomly on the map. As the three adversaries distribute in separate positions, the agents naturally form three groups to chase different adversaries.
    \item After several time steps, in Fig.~\ref{figure:tag_illustration}-b, two adversaries gather at the bottom left of the map. Note that, in this game, each agent can only observe objects within a small circular range. Only agent $1$ can observe the two adversaries simultaneously, which makes it connect to the majority of the rest of the agents and helps them recognize the position of the adversaries.
    \item Finally, as illustrated by Fig.~\ref{figure:tag_illustration}-c, the bottom adversary is surrounded by multiple agents, yielding an opportunity for the agents to win considerable rewards, which makes the collaboration structure concentrate around the adversary.
\end{enumerate}
As presented above, the graph structures learned by SOP-CG can learn effective coordination patterns in Tag, which can be explained by the oracle collaboration strategy of human experts. It demonstrates the ability of our approach to organize coordination relations.
\section{Conclusion}
This paper introduces SOP-CG, a novel multi-agent graph-based method that guarantees the polynomial-time greedy policy execution with expressive function representation of a structured graph class. To enable end-to-end learning, SOP-CG introduces an imaginary agent to dynamically select the state-dependent graph and incorporates it into a unified Bellman optimality equation. We demonstrate that SOP-CG can learn interpretable graph topologies and outperform state-of-the-art baselines on several challenging tasks. Although the graph class is exponentially large, we conduct two specific polynomial-time graph class instances to achieve impressive performance. Designing effective graph exploration methods to deal with larger graph classes will be an interesting future direction.

\section{Acknowledgement}
This work is supported in part by Science and Technology Innovation 2030 – ``New Generation Artificial Intelligence" Major Project (No. 2018AAA0100904) and National Natural Science Foundation of China (62176135).

\newpage

\bibliography{ref}

\begin{thebibliography}{47}
\providecommand{\natexlab}[1]{#1}
\providecommand{\url}[1]{\texttt{#1}}
\expandafter\ifx\csname urlstyle\endcsname\relax
  \providecommand{\doi}[1]{doi: #1}\else
  \providecommand{\doi}{doi: \begingroup \urlstyle{rm}\Url}\fi

\bibitem[Alonso-Mora et~al.(2017)Alonso-Mora, Baker, and Rus]{alonso2017multi}
Alonso-Mora, J., Baker, S., and Rus, D.
\newblock Multi-robot formation control and object transport in dynamic
  environments via constrained optimization.
\newblock \emph{The International Journal of Robotics Research}, 36\penalty0
  (9):\penalty0 1000--1021, 2017.

\bibitem[B{\"o}hmer et~al.(2020)B{\"o}hmer, Kurin, and
  Whiteson]{bohmer2020deep}
B{\"o}hmer, W., Kurin, V., and Whiteson, S.
\newblock Deep coordination graphs.
\newblock In \emph{International Conference on Machine Learning}, pp.\
  980--991. PMLR, 2020.

\bibitem[Castellini et~al.(2019)Castellini, Oliehoek, Savani, and
  Whiteson]{castellini2019representational}
Castellini, J., Oliehoek, F.~A., Savani, R., and Whiteson, S.
\newblock The representational capacity of action-value networks for
  multi-agent reinforcement learning.
\newblock In \emph{AAMAS 2019: The 18th International Conference on Autonomous
  Agents and MultiAgent Systems}, pp.\  1862--1864. International Foundation
  for Autonomous Agents and Multiagent Systems (IFAAMAS), 2019.

\bibitem[Chan(2016)]{chan2016approximation}
Chan, S.~O.
\newblock Approximation resistance from pairwise-independent subgroups.
\newblock \emph{Journal of the ACM (JACM)}, 63\penalty0 (3):\penalty0 1--32,
  2016.

\bibitem[Cho et~al.(2014)Cho, Van~Merri{\"e}nboer, Gulcehre, Bahdanau,
  Bougares, Schwenk, and Bengio]{cho2014learning}
Cho, K., Van~Merri{\"e}nboer, B., Gulcehre, C., Bahdanau, D., Bougares, F.,
  Schwenk, H., and Bengio, Y.
\newblock Learning phrase representations using rnn encoder-decoder for
  statistical machine translation.
\newblock \emph{arXiv preprint arXiv:1406.1078}, 2014.

\bibitem[Dagum \& Luby(1993)Dagum and Luby]{dagum1993approximating}
Dagum, P. and Luby, M.
\newblock Approximating probabilistic inference in bayesian belief networks is
  np-hard.
\newblock \emph{Artificial intelligence}, 60\penalty0 (1):\penalty0 141--153,
  1993.

\bibitem[Das et~al.(2019)Das, Gervet, Romoff, Batra, Parikh, Rabbat, and
  Pineau]{das2019tarmac}
Das, A., Gervet, T., Romoff, J., Batra, D., Parikh, D., Rabbat, M., and Pineau,
  J.
\newblock Tarmac: Targeted multi-agent communication.
\newblock In \emph{International Conference on Machine Learning}, pp.\
  1538--1546, 2019.

\bibitem[Edmonds(1965)]{edmonds1965paths}
Edmonds, J.
\newblock Paths, trees, and flowers.
\newblock \emph{Canadian Journal of mathematics}, 17:\penalty0 449--467, 1965.

\bibitem[Fioretto et~al.(2018)Fioretto, Pontelli, and
  Yeoh]{fioretto2018distributed}
Fioretto, F., Pontelli, E., and Yeoh, W.
\newblock Distributed constraint optimization problems and applications: A
  survey.
\newblock \emph{Journal of Artificial Intelligence Research}, 61:\penalty0
  623--698, 2018.

\bibitem[Foerster et~al.(2017)Foerster, Nardelli, Farquhar, Afouras, Torr,
  Kohli, and Whiteson]{foerster2017stabilising}
Foerster, J., Nardelli, N., Farquhar, G., Afouras, T., Torr, P.~H., Kohli, P.,
  and Whiteson, S.
\newblock Stabilising experience replay for deep multi-agent reinforcement
  learning.
\newblock In \emph{International conference on machine learning}, pp.\
  1146--1155. PMLR, 2017.

\bibitem[Foerster et~al.(2018)Foerster, Farquhar, Afouras, Nardelli, and
  Whiteson]{foerster2018counterfactual}
Foerster, J.~N., Farquhar, G., Afouras, T., Nardelli, N., and Whiteson, S.
\newblock Counterfactual multi-agent policy gradients.
\newblock In \emph{Thirty-Second AAAI Conference on Artificial Intelligence},
  2018.

\bibitem[Guestrin et~al.(2001)Guestrin, Koller, and
  Parr]{guestrin2001multiagent}
Guestrin, C., Koller, D., and Parr, R.
\newblock Multiagent planning with factored mdps.
\newblock In \emph{NIPS}, volume~1, pp.\  1523--1530, 2001.

\bibitem[Guestrin et~al.(2002{\natexlab{a}})Guestrin, Lagoudakis, and
  Parr]{guestrin2002coordinated}
Guestrin, C., Lagoudakis, M., and Parr, R.
\newblock Coordinated reinforcement learning.
\newblock In \emph{ICML}, volume~2, pp.\  227--234. Citeseer,
  2002{\natexlab{a}}.

\bibitem[Guestrin et~al.(2002{\natexlab{b}})Guestrin, Venkataraman, and
  Koller]{guestrin2002context}
Guestrin, C., Venkataraman, S., and Koller, D.
\newblock Context-specific multiagent coordination and planning with factored
  mdps.
\newblock In \emph{AAAI/IAAI}, pp.\  253--259, 2002{\natexlab{b}}.

\bibitem[Hasselt(2010)]{hasselt2010double}
Hasselt, H.
\newblock Double q-learning.
\newblock \emph{Advances in neural information processing systems},
  23:\penalty0 2613--2621, 2010.

\bibitem[Jiang \& Lu(2021)Jiang and Lu]{jiang2021emergence}
Jiang, J. and Lu, Z.
\newblock The emergence of individuality.
\newblock In \emph{International Conference on Machine Learning}, pp.\
  4992--5001. PMLR, 2021.

\bibitem[Kok \& Vlassis(2006)Kok and Vlassis]{kok2006collaborative}
Kok, J.~R. and Vlassis, N.
\newblock Collaborative multiagent reinforcement learning by payoff
  propagation.
\newblock \emph{Journal of Machine Learning Research}, 7:\penalty0 1789--1828,
  2006.

\bibitem[Kraemer \& Banerjee(2016)Kraemer and Banerjee]{kraemer2016multi}
Kraemer, L. and Banerjee, B.
\newblock Multi-agent reinforcement learning as a rehearsal for decentralized
  planning.
\newblock \emph{Neurocomputing}, 190:\penalty0 82--94, 2016.

\bibitem[Li et~al.(2020)Li, Gupta, Morales, Allen, and
  Kochenderfer]{li2020deep}
Li, S., Gupta, J.~K., Morales, P., Allen, R., and Kochenderfer, M.~J.
\newblock Deep implicit coordination graphs for multi-agent reinforcement
  learning.
\newblock \emph{arXiv preprint arXiv:2006.11438}, 2020.

\bibitem[Lowe et~al.(2017)Lowe, Wu, Tamar, Harb, Abbeel, and
  Mordatch]{lowe2017multi}
Lowe, R., Wu, Y., Tamar, A., Harb, J., Abbeel, P., and Mordatch, I.
\newblock Multi-agent actor-critic for mixed cooperative-competitive
  environments.
\newblock \emph{Neural Information Processing Systems (NIPS)}, 2017.

\bibitem[Manurangsi et~al.(2015)Manurangsi, Nakkiran, and
  Trevisan]{manurangsi2015near}
Manurangsi, P., Nakkiran, P., and Trevisan, L.
\newblock Near-optimal ugc-hardness of approximating max k-csp\_r.
\newblock \emph{arXiv preprint arXiv:1511.06558}, 2015.

\bibitem[Oliehoek et~al.(2016)Oliehoek, Amato, et~al.]{oliehoek2016concise}
Oliehoek, F.~A., Amato, C., et~al.
\newblock \emph{A concise introduction to decentralized POMDPs}, volume~1.
\newblock Springer, 2016.

\bibitem[OroojlooyJadid \& Hajinezhad(2019)OroojlooyJadid and
  Hajinezhad]{oroojlooyjadid2019review}
OroojlooyJadid, A. and Hajinezhad, D.
\newblock A review of cooperative multi-agent deep reinforcement learning.
\newblock \emph{arXiv preprint arXiv:1908.03963}, 2019.

\bibitem[Panait et~al.(2006)Panait, Luke, and Wiegand]{panait2006biasing}
Panait, L., Luke, S., and Wiegand, R.~P.
\newblock Biasing coevolutionary search for optimal multiagent behaviors.
\newblock \emph{IEEE Transactions on Evolutionary Computation}, 10\penalty0
  (6):\penalty0 629--645, 2006.

\bibitem[Park \& Darwiche(2004)Park and Darwiche]{park2004complexity}
Park, J.~D. and Darwiche, A.
\newblock Complexity results and approximation strategies for map explanations.
\newblock \emph{Journal of Artificial Intelligence Research}, 21:\penalty0
  101--133, 2004.

\bibitem[Pearl(1988)]{pearl1988probabilistic}
Pearl, J.
\newblock \emph{Probabilistic Reasoning in Intelligent Systems: Networks of
  Plausible Inference}.
\newblock Morgan Kaufmann, 1988.

\bibitem[Rashid et~al.(2020{\natexlab{a}})Rashid, Farquhar, Peng, and
  Whiteson]{rashid2020weighted}
Rashid, T., Farquhar, G., Peng, B., and Whiteson, S.
\newblock Weighted qmix: Expanding monotonic value function factorisation for
  deep multi- agent reinforcement learning.
\newblock \emph{NeurIPS Proceedings 2020}, 2020{\natexlab{a}}.

\bibitem[Rashid et~al.(2020{\natexlab{b}})Rashid, Samvelyan, De~Witt, Farquhar,
  Foerster, and Whiteson]{rashid2020monotonic}
Rashid, T., Samvelyan, M., De~Witt, C.~S., Farquhar, G., Foerster, J., and
  Whiteson, S.
\newblock Monotonic value function factorisation for deep multi-agent
  reinforcement learning.
\newblock \emph{arXiv preprint arXiv:2003.08839}, 2020{\natexlab{b}}.

\bibitem[Samvelyan et~al.(2019)Samvelyan, Rashid, de~Witt, Farquhar, Nardelli,
  Rudner, Hung, Torr, Foerster, and Whiteson]{samvelyan19smac}
Samvelyan, M., Rashid, T., de~Witt, C.~S., Farquhar, G., Nardelli, N., Rudner,
  T. G.~J., Hung, C.-M., Torr, P. H.~S., Foerster, J., and Whiteson, S.
\newblock {The} {StarCraft} {Multi}-{Agent} {Challenge}.
\newblock \emph{CoRR}, abs/1902.04043, 2019.

\bibitem[Singh et~al.(2018)Singh, Jain, and Sukhbaatar]{singh2018learning}
Singh, A., Jain, T., and Sukhbaatar, S.
\newblock Learning when to communicate at scale in multiagent cooperative and
  competitive tasks.
\newblock \emph{arXiv preprint arXiv:1812.09755}, 2018.

\bibitem[Son et~al.(2019)Son, Kim, Kang, Hostallero, and Yi]{son2019qtran}
Son, K., Kim, D., Kang, W.~J., Hostallero, D.~E., and Yi, Y.
\newblock Qtran: Learning to factorize with transformation for cooperative
  multi-agent reinforcement learning.
\newblock In \emph{International Conference on Machine Learning}, pp.\
  5887--5896, 2019.

\bibitem[Stranders et~al.(2009)Stranders, Farinelli, Rogers, and
  Jennings]{stranders2009decentralised}
Stranders, R., Farinelli, A., Rogers, A., and Jennings, N.~R.
\newblock Decentralised coordination of mobile sensors using the max-sum
  algorithm.
\newblock In \emph{Twenty-First International Joint Conference on Artificial
  Intelligence}, 2009.

\bibitem[Sunehag et~al.(2018)Sunehag, Lever, Gruslys, Czarnecki, Zambaldi,
  Jaderberg, Lanctot, Sonnerat, Leibo, Tuyls, et~al.]{sunehag2018value}
Sunehag, P., Lever, G., Gruslys, A., Czarnecki, W.~M., Zambaldi, V., Jaderberg,
  M., Lanctot, M., Sonnerat, N., Leibo, J.~Z., Tuyls, K., et~al.
\newblock Value-decomposition networks for cooperative multi-agent learning
  based on team reward.
\newblock In \emph{Proceedings of the 17th International Conference on
  Autonomous Agents and MultiAgent Systems}, pp.\  2085--2087, 2018.

\bibitem[Tan(1993)]{tan1993multi}
Tan, M.
\newblock Multi-agent reinforcement learning: Independent vs. cooperative
  agents.
\newblock In \emph{Proceedings of the tenth international conference on machine
  learning}, pp.\  330--337, 1993.

\bibitem[Van~der Pol \& Oliehoek(2016)Van~der Pol and
  Oliehoek]{van2016coordinated}
Van~der Pol, E. and Oliehoek, F.~A.
\newblock Coordinated deep reinforcement learners for traffic light control.
\newblock \emph{Proceedings of Learning, Inference and Control of Multi-Agent
  Systems (at NIPS 2016)}, 2016.

\bibitem[Wang et~al.(2021{\natexlab{a}})Wang, Ren, Liu, Yang, and
  Zhang]{wang2020QPLEX}
Wang, J., Ren, Z., Liu, T., Yang, Y., and Zhang, C.
\newblock Qplex: Duplex dueling multi-agent q-learning.
\newblock In \emph{International Conference on Learning Representations},
  2021{\natexlab{a}}.

\bibitem[Wang et~al.(2020{\natexlab{a}})Wang, Dong, Lesser, and
  Zhang]{wang2020multi}
Wang, T., Dong, H., Lesser, V., and Zhang, C.
\newblock Multi-agent reinforcement learning with emergent roles.
\newblock In \emph{International Conference on Machine Learning},
  2020{\natexlab{a}}.

\bibitem[Wang et~al.(2020{\natexlab{b}})Wang, Gupta, Mahajan, Peng, Whiteson,
  and Zhang]{wang2020rode}
Wang, T., Gupta, T., Mahajan, A., Peng, B., Whiteson, S., and Zhang, C.
\newblock Rode: Learning roles to decompose multi-agent tasks.
\newblock \emph{arXiv preprint arXiv:2010.01523}, 2020{\natexlab{b}}.

\bibitem[Wang et~al.(2020{\natexlab{c}})Wang, Wang, Zheng, and
  Zhang]{wang2019learning}
Wang, T., Wang, J., Zheng, C., and Zhang, C.
\newblock Learning nearly decomposable value functions via communication
  minimization.
\newblock In \emph{International Conference on Learning Representations},
  2020{\natexlab{c}}.

\bibitem[Wang et~al.(2021{\natexlab{b}})Wang, Zeng, Dong, Yang, Yu, and
  Zhang]{wang2021context}
Wang, T., Zeng, L., Dong, W., Yang, Q., Yu, Y., and Zhang, C.
\newblock Context-aware sparse deep coordination graphs.
\newblock \emph{arXiv preprint arXiv:2106.02886}, 2021{\natexlab{b}}.

\bibitem[Wang et~al.(2020{\natexlab{d}})Wang, Han, Wang, Dong, and
  Zhang]{wang2020off}
Wang, Y., Han, B., Wang, T., Dong, H., and Zhang, C.
\newblock Off-policy multi-agent decomposed policy gradients.
\newblock \emph{arXiv preprint arXiv:2007.12322}, 2020{\natexlab{d}}.

\bibitem[Wei \& Luke(2016)Wei and Luke]{wei2016lenient}
Wei, E. and Luke, S.
\newblock Lenient learning in independent-learner stochastic cooperative games.
\newblock \emph{The Journal of Machine Learning Research}, 17\penalty0
  (1):\penalty0 2914--2955, 2016.

\bibitem[Wen et~al.(2019)Wen, Yang, Luo, Wang, and Pan]{wen2019probabilistic}
Wen, Y., Yang, Y., Luo, R., Wang, J., and Pan, W.
\newblock Probabilistic recursive reasoning for multi-agent reinforcement
  learning.
\newblock \emph{arXiv preprint arXiv:1901.09207}, 2019.

\bibitem[Ye et~al.(2015)Ye, Zhang, and Yang]{ye2015multi}
Ye, D., Zhang, M., and Yang, Y.
\newblock A multi-agent framework for packet routing in wireless sensor
  networks.
\newblock \emph{Sensors}, 15\penalty0 (5):\penalty0 10026--10047, 2015.

\bibitem[Yoon et~al.(2020)Yoon, Song, Shin, and Yi]{yoon2020much}
Yoon, S.-e., Song, H., Shin, K., and Yi, Y.
\newblock How much and when do we need higher-order information in hypergraphs?
  a case study on hyperedge prediction.
\newblock In \emph{Proceedings of The Web Conference 2020}, pp.\  2627--2633,
  2020.

\bibitem[Zhang \& Lesser(2011)Zhang and Lesser]{zhang2011coordinated}
Zhang, C. and Lesser, V.
\newblock Coordinated multi-agent reinforcement learning in networked
  distributed pomdps.
\newblock In \emph{Twenty-Fifth AAAI Conference on Artificial Intelligence},
  2011.

\bibitem[Zhang \& Lesser(2013)Zhang and Lesser]{zhang2013coordinating}
Zhang, C. and Lesser, V.
\newblock Coordinating multi-agent reinforcement learning with limited
  communication.
\newblock In \emph{Proceedings of the 2013 international conference on
  Autonomous agents and multi-agent systems}, pp.\  1101--1108, 2013.

\end{thebibliography}
\bibliographystyle{icml2022}

\newpage
\appendix
\onecolumn

\section{Experiment Details of the Motivating Example}
\label{appendix:dcop_test}

\paragraph{Algorithms.} To get the correct result of DCOP, we use a brute force with complexity $O(|A|^n)$, while we choose Max-sum algorithm (has an alternative implementation called Max-plus)~\citep{pearl1988probabilistic}, which is commonly used by most of the coordination graph learning methods \citep{zhang2013coordinating, bohmer2020deep, wang2021context} to test the accuracy and relative Q error. The details of the algorithm are presented  below~\citep{zhang2013coordinating}.

Max-sum constructs a bipartite graph $\mathcal{G}_m = \langle \mathcal{V}_a, \mathcal{V}_q, \mathcal{E}_m \rangle$ according the coordination graph $\langle \mathcal{V}, \mathcal{E} \rangle$. Each node $v \in \mathcal{V}_m$ represents an agent who needs to do action selection, and each node $u \in \mathcal{V}_q$ represents a (hyper-)edge function. Edges in $\mathcal{E}_m$ connect $u$ with the corresponding agent nodes. Max-sum algorithm will do multi-round message passing on this graph. The number of iterations is usually set smaller than $10$ in previous work, while we set it $100$ in this experiment.

Message passing on this bipartite graph starts with sending messages from node $v \in \mathcal{V}_a$ to node $u \in \mathcal{V}_q$:

\begin{equation}
    m_{v \to u}(a_i) = \sum_{h\in\mathcal{F}_v \backslash u} m_{h \to v} (a_i) + c_{vu}
\end{equation}

where $\mathcal{F}_v$ represents the node connected to $v$, and $c_{vu}$ is the normalizing factor preventing the value of messages from growing arbitrarily large. The message from node $u$ to $v$ is:

\begin{equation}
    m_{u \to v}(a_i) = \max_{\bm{a}_u \backslash a_v} \left [ f(\bm{a}_u) + \sum_{h \in \mathcal{V}_u \backslash v} m_{h \to u}(a_h)  \right]
\end{equation}

where $\mathcal{V}_{u}$ is the set of nodes connected to node $u$, $\bm{a}_{u} = \left\{a_{h}| h \in \mathcal{V}_{u}\right\}$, $\bm{a}_{u} \backslash a_{v} = \left\{a_{h}| h \in \mathcal{V}_{u} \backslash \{v\}\right\}$, and $f$ represents the value function of the (hyper-)edge. After iterations of message passing, each agent $v$ can find its optimal action by calculating $a_{v}^{*}=\operatorname{argmax}_{a_{v}}\sum_{h \in \mathcal{F}_{v}} m_{h \rightarrow v}\left(a_{v}\right)$.

\paragraph{Experiment setup.} We take $1000$ random seeds to test the problem. We fix the number of actions $3$ and move the number of agents from $2$ to $18$. For each setting, we generate a fully connected graph with each value a uniformly random number in $\{-1,0,1\}$ along with a noise drawn from the normal distribution.


\section{Proof of Proposition~\ref{proposition:dcop_complexity}}
\label{appendix:proof}
In this section, we provide the proof of Proposition \ref{proposition:dcop_complexity} (i). Proposition \ref{proposition:dcop_complexity} (ii) is given in the literature \cite{fioretto2018distributed}.

We will reduce Max 2CSP-R \citep{manurangsi2015near} to the DCOPs induced by fully connected graphs. Max 2CSP-R is a constraint satisfaction problem with two variables per constraint, where each variable can take values from a finite set $\Sigma$ of size R, and the goal is to find an assignment to variables that maximizes the number of satisfied constraints.

For each variable in the Max 2CSP-R problem, we create an agent in the coordination graph with $R$ actions. Furthermore, for each constraint in the Max 2CSP-R problem, we create an edge in the coordination graph, connecting the two agents of the variables, with a function value $1$ only when the constraint is satisfied and otherwise $0$. If we can approximate the induced DCOP problem within a factor of $c\frac{\log R}{\sqrt{R}}$, we can  distinguish the $(1-\epsilon)$-satisfiable Max 2CSP-R instances and the instances in which at most a $c \frac{\log R}{\sqrt{R}}$ fraction of constraints are satisfiable. \citet{chan2016approximation} proves that there is a constant $c$ such that for every  sufficiently large R and any $\epsilon>0$, it is NP-hard to distinguish $(1 - \epsilon)$-satisfiable instances of Max 2CSP-R from instances in which at most a $c \frac{\log R}{\sqrt{R}}$ fraction of constraints are satisfiable. Therefore, it is NP-hard to Approximate the induced DCOPs of fully connected coordination graphs within a factor of $O\left(\frac{\log A}{\sqrt A}\right)$. Results given by \citet{manurangsi2015near} also show that we can get a tighter approximation hardness bound of $O\left(\frac{\log A}{A}\right)$ if assuming the unique games conjecture.

\section{Implementation tricks of Temporal Difference Learning}
\label{appendix:graph_relabel}
As studied by previous work \cite{hasselt2010double}, vanilla Q-learning already has a tendency to overestimate, since the TD target takes a max operator over a set of estimated action-values. Notice that our framework introduces an additional max operator over graphs in class $\mathcal{G}$, which will probably induce extra upward bias. To attenuate this problem, we use a graph relabeling technique to control the effects of overestimation errors. In our implementation, we modify the vanilla TD loss defined in Eq.~\eqref{equation:td_loss} to the following form: 
\begin{equation}
\mathcal{L}_{cg}(\bm{\theta})=\mathop{\mathbb{E}}_{(\boldsymbol{\tau},\boldsymbol{a},r,\boldsymbol{\tau}')\sim \mathcal{D}}\left[\left(y_{cg}-\max_{G} Q(\boldsymbol{\tau},\boldsymbol{a};G;\bm{\theta})\right)^2\right]
\label{equation:modified_td_loss}
\end{equation}
where $y_{cg}$ is the one-step TD target defined in Section~\ref{method:framework} and the current value no longer takes the graph which is stored in the replay buffer with this transition. As $G$ is the graph with largest estimated value, this loss achieves a soft constraint for the following target:
\begin{equation}
\forall G, Q(\boldsymbol{\tau},\boldsymbol{a};G;\bm{\theta}) \leq y_{cg}
\label{equation:constraint}
\end{equation}
which helps our method avoid overestimation.

\begin{figure}[t]
    \centering
    \includegraphics[width=0.65\linewidth]{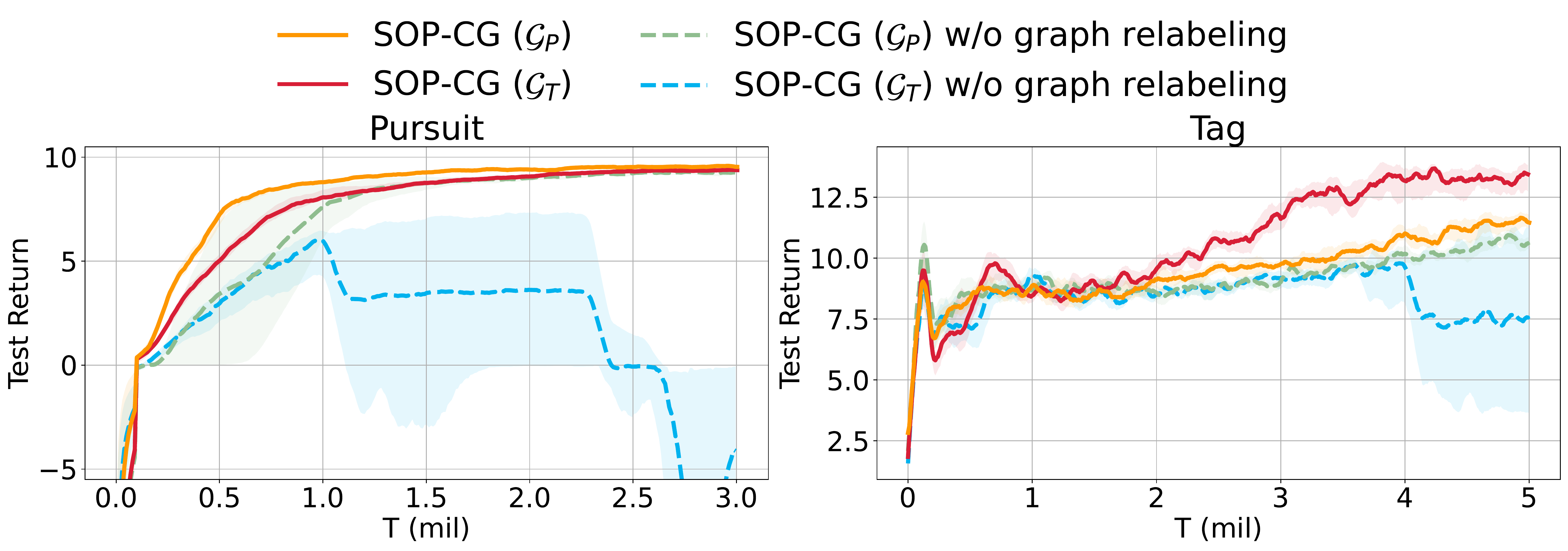}
    \vspace{-0.1in}
    \caption{Ablation studies about the graph relabeling technique.}
    \label{figure:ablation-relabel}
\end{figure}

We conduct ablation studies on the tasks Pursuit and Tag to demonstrate the necessity of this component. The results illustrated in Fig.~\ref{figure:ablation-relabel}. SOP-CG w/o graph relabeling is more likely to stick in suboptimal strategies and may even collapse on Pursuit. As analyzed before, these phenomenons are due to the dramatic overestimation caused by the additional max operator over graphs. By contrast, the performance of SOP-CG is significantly improved, highlighting the effectiveness of graph relabeling.

\section{Computations with Classes $\mathcal{G}_P$ and $\mathcal{G}_T$}
\label{appendix:algorithm}

Here we present the details of algorithms we used to select graph $G^{(t)}$ from the graph class $\mathcal{G}$. 

\paragraph{When $\mathcal{G} = \mathcal{G}_P$ (pairwise grouping).} In this case, $n$ agents are pairwise matched by $\lfloor n/2 \rfloor$ edges. One agent will be isolated if $n$ is an odd number. With such a coordination graph, each agent coordinates with exactly one agent, except for the isolated one.

If node $i$ is matched with node $j$, they will contribute at most
\begin{align}
    \max_{a_i,a_j} \left[ q_i(\tau_i,a_i)+q_j(\tau_j,a_j)+q_{ij}(\tau_i,\tau_j,a_i,a_j)\right]
\end{align}
to the joint value without affecting the action selection of other agents.

Hence, we construct a undirected weighted graph $G = \langle \mathcal{V}, \mathcal{E}_w \rangle$, where $\mathcal{V}$ is the vertex set, and 
\begin{align}
    \mathcal{E}_w = \left\{\left(i,j, w(i,j) = \max_{a_i,a_j} \left[ q_i(\tau_i,a_i)+q_j(\tau_j,a_j)+q_{ij}(\tau_i,\tau_j,a_i,a_j)\right]\right) \big| \forall (i,j) \in \mathcal{E}\right\}.
\end{align} 
A matching on such graph is a function $mate(v)$ where $mate(mate(v)) = v$ for all $v\in\mathcal{V}$ with at most one $v$ satisfies that $mate(v) = v$. The weight of the matching is defined as $\frac{1}{2}\sum_v w(v, mate(v)) $. Our goal is to find the maximum weighted matching on this graph.

This problem can be solved by the blossom algorithm~\citep{edmonds1965paths} in $O(n^3)$ time. The brief steps of this algorithm are to iteratively find augmentation paths on the graph and contract the odd-cycle, which is called blossom in this algorithm.

\paragraph{When $\mathcal{G}=\mathcal{G}_T$ (tree organization).} In this case, the graph is an acyclic graph with $n-1$ edges, and all agents form a connected component. The distributed constraint optimization problem (DCOP) on such graphs can always be solved in the polynomial time~\citep{fioretto2018distributed}.

Choosing the best graph $G^{(t)}$ in $\mathcal{G}_T$ is non-trivial. Therefore, we present a greedy algorithm to select the graph from the graph class. The high level idea is to add edges greedily until all agents are connected. At each iteration, we add an edge across two different connected components to the coordination graph, which optimizes the increment of the maximal joint value. For an edge set $E \subseteq \mathcal{E}$, we
define the value of a set $f(E)$ to be the result of DCOP on graph $\langle \mathcal{V}, E \rangle$. Initially we have $E$ is an empty set. We do the iteration $n - 1$ times. In each iteration, we find 
\begin{align}
    e=\mathop{\arg\max}_{e \in \mathcal{E}, E \cup \{e\} \text{~is~acyclic}} f (E \cup \{e\}),
\end{align}
and add this edge to the current graph, i.e., $E \gets E \cup \{e\}$. Once a graph from $\mathcal{G}_T$ is selected, the joint action selection can be computed via dynamic programming.

\paragraph{Time complexity.} We summarize the time complexity (in worst case) of each component in Table~\ref{table:time_complexity}. For SOP-CG ($\mathcal{G}_T$), we use pruning techniques in our implementation, so the real-time cost of selecting graph $G^{(t)}$ is usually much smaller than the complexity reported in Table~\ref{table:time_complexity}. The detailed training time cost of SOP-CG and DCG can be found in Appendix~\ref{appendix:hyperparameter}. 

\begin{table}[h!]
    \centering
    \begin{tabular}{crcrcr}
        \toprule
        \multicolumn{2}{c}{Algorithm} & \multicolumn{2}{c}{\makecell[c]{Select graph $G^{(t)}$}} & \multicolumn{2}{c}{\makecell[c]{Select actions on a given graph}} \\
        \cmidrule(lr){1-2} \cmidrule(lr){3-4} \cmidrule(lr){5-6}
        \multicolumn{2}{c}{SOP-CG ($\mathcal{G}_P$)} & \multicolumn{2}{c}{$O(n^3)$} & \multicolumn{2}{c}{$O(nA^2)$} \\
        \cmidrule(lr){1-2} \cmidrule(lr){3-4} \cmidrule(lr){5-6}
        \multicolumn{2}{c}{SOP-CG ($\mathcal{G}_T$)} & \multicolumn{2}{c}{$O(n^3A^2)$ (heuristic)} & \multicolumn{2}{c}{$O(nA^2)$} \\
        \cmidrule(lr){1-2} \cmidrule(lr){3-4} \cmidrule(lr){5-6}
        \multicolumn{2}{c}{DCG} & \multicolumn{2}{c}{N/A} & \multicolumn{2}{c}{$O(kn^2A^2)$ (heuristic)} \\
        \bottomrule
    \end{tabular}
    \caption{Time complexity of SOP-CG and DCG. $n$ is the number of agents and $A=|\bigcup_{i=1}^n A^i|$. For DCG, $k$ is the number of iterations in max-sum.}
    \label{table:time_complexity}
\end{table}

\section{Settings and Results of the Illustrative Example}
\label{appendix:illustrative}

\paragraph{Experiment Setting.}

In this illustrative example, each agent performs independent $\epsilon$-greedy exploration, with $\epsilon$ anneals linearly from $1.0$ to $0.05$ over $100000$ time-steps. We set $\gamma = 0.99$. The replay buffer consists of the last $500$ episodes, from which we uniformly sample a batch of size 32 for training.

The group allocation is drawn from the uniform distribution, with each time only one group having three agents can play action B. The group id is one-hot encoded. Each agent receives the group id and its available actions as input.

\begin{wrapfigure}[8]{r}{0.33\linewidth}
    \centering
    \vspace{-1cm}
    \includegraphics[width=\linewidth]{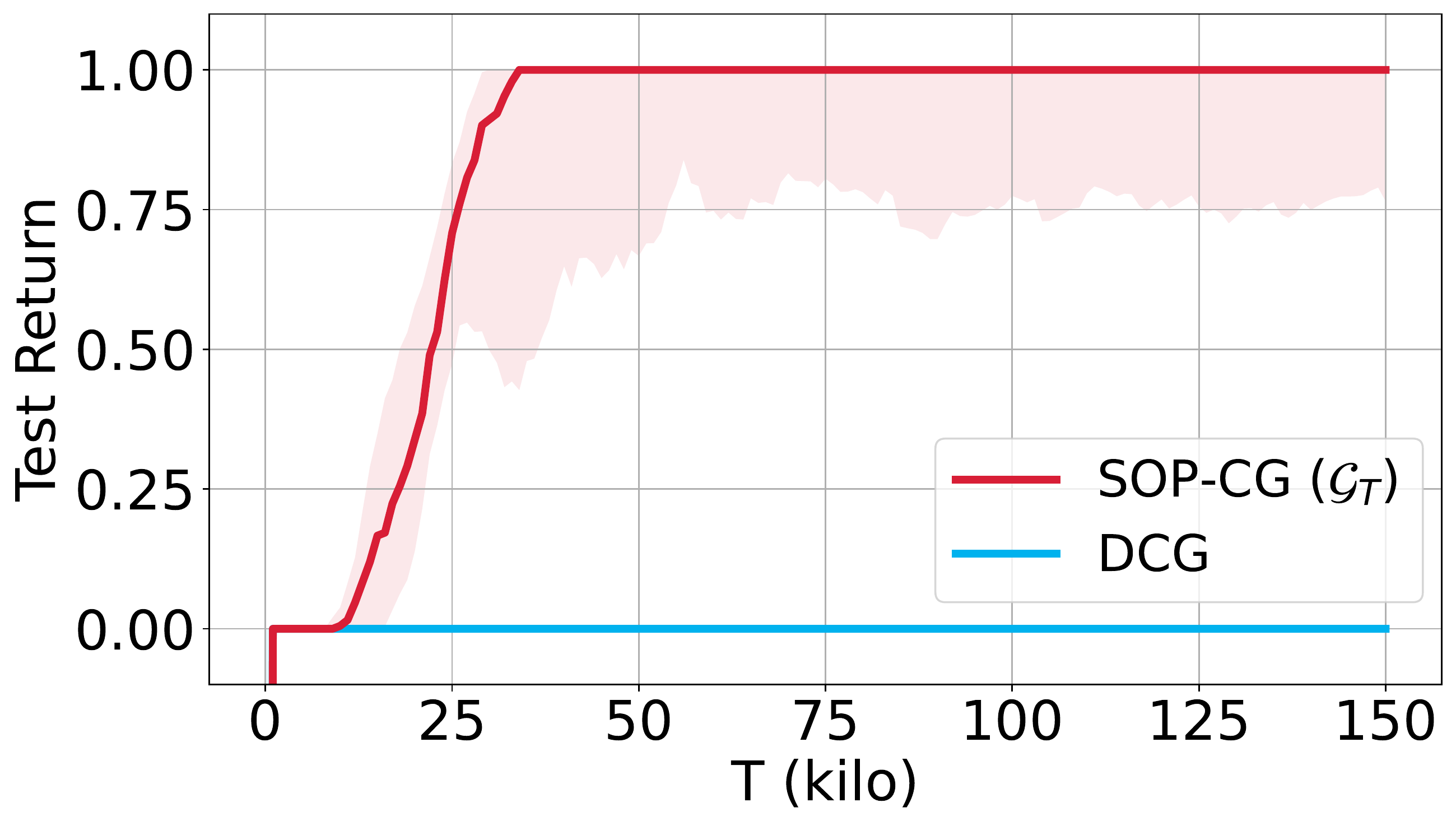}
    \vspace{-2.3em}
    \caption{Extension of the illustrative example with 5 groups and each group 4 agents.}
    \label{figure:toy-4tuple}
\end{wrapfigure}

\paragraph{Detailed results.}

\begin{figure*}[htbp]
    \centering
    \includegraphics[width=\linewidth]{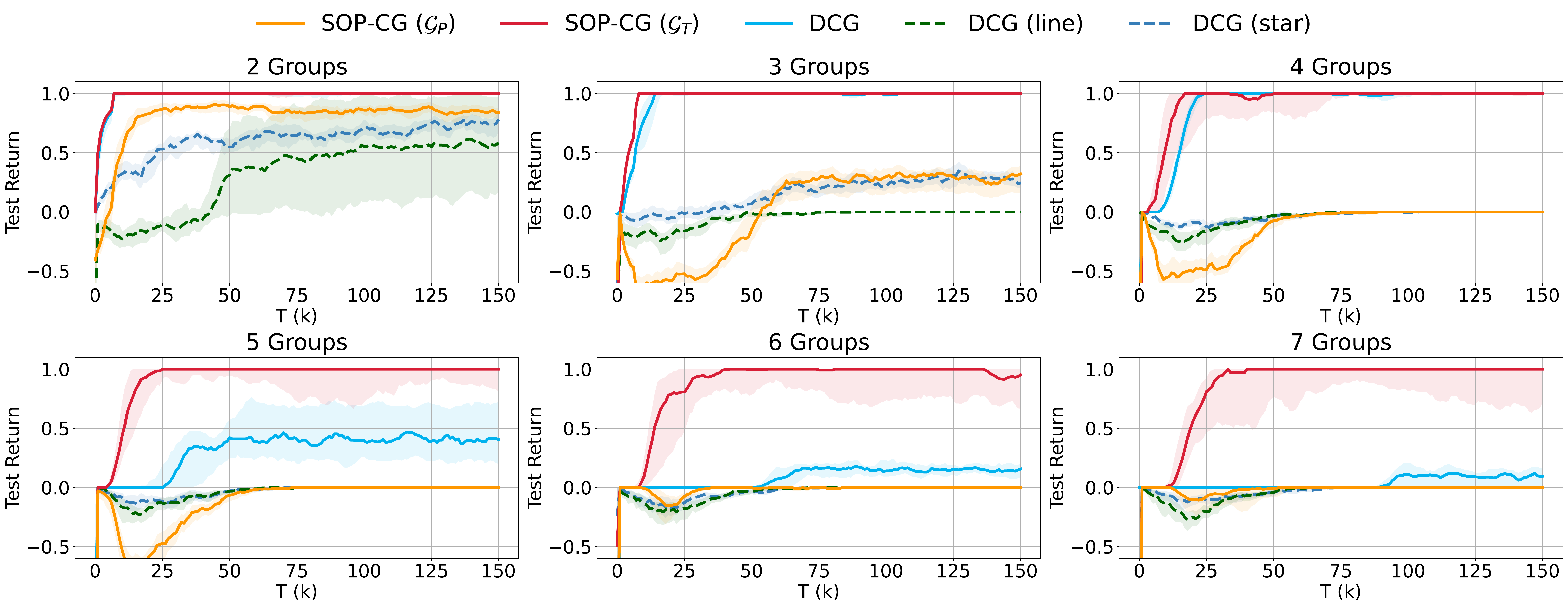}
    \vspace{-0.3in}
    \caption{Learning curves on the illustrative example.}
    \label{figure:toy-full}
\end{figure*}

Fig.~\ref{figure:toy-full} shows the full result of our method and DCG (line, star and full) on this environment. 

We also test an extended version of the illustrative example. In this version, each group has four agents. Choosing action B will still yield a cost of 0.5. If all
agents in the same group take action B, they will receive a
global reward of 3. Therefore, this version requires coordination among 4-tuples of agents. As shown in Figure~\ref{figure:toy-4tuple}, SOP-CG ($\mathcal{G}_{T}$) still works well. 

\section{Experiment Settings and Hyperparameters}
\label{appendix:hyperparameter}

\subsection{Settings of Experiment Tasks}
MACO \citep{wang2021context} is a benchmark including 6 classic tasks: Aloha, Pursuit, Hallway, Sensor, Gather and Disperse. For the task Pursuit, MACO uses the version with 10 predators and 5 prey. In this paper, we use a harder version with 20 predators and 10 prey. For the other tasks, we test the vanilla version in MACO. The number of agents and actions are listed in Table~\ref{table:maco_number}.

\begin{table}[h!]
    \centering
    \begin{tabular}{|c|c|c|c|c|c|c|}
      \hline
       & Aloha & Pursuit & Hallway & Sensor & Gather & Disperse \\ \hline
      \# agents & 10 & 20 & 12 & 15 & 5 & 12 \\ \hline
      \# actions & 2 & 9 & 3 & 9 & 5 & 4 \\ \hline
    \end{tabular}
    \caption{Number of agents and actions in the tasks from MACO.}
    \label{table:maco_number}
\end{table}

Tag is a challenging task on the particle world  \citep{lowe2017multi}. On the map with 3 randomly generated obstacles, 10 slower agents chase 3 faster adversaries. For each collision between an agent and an adversary, the agents receive a reward while the adversaries are punished. Each agent has 5 actions to control its movement. An agent or adversary can observe the units in a restricted circular area that is centered at its position. We train the agents by a specific algorithm, while the adversaries are concurrently trained by VDN \citep{sunehag2018value}.

SMAC \citep{samvelyan19smac} is a popular benchmark based on the strategy game StarCraft II. In this paper, we benchmark our algorithm with StarCraft version of 2.4.6.2.69232, which is used by the SMAC paper.

\subsection{Hyperparameters}
A shared GRU \citep{cho2014learning} is used to process sequential inputs and then output the encoded observation history for each agent. A following fully-connected layer converts the observation histories to the individual utility function of each agent. The pairwise payoff function is computed by another multi-layer perceptron, which takes the concatenation of two agents' observation histories as input. Agents share parameters of the network for computing individual utility, and the parameters of the payoff network are also shared among different agent pairs.

All tasks in this paper use a discount factor $\gamma=0.99$. We use $\epsilon$-greedy exploration, and $\epsilon$ anneals linearly from 1.0 to 0.05 over 50000 time-steps. We use an RMSProp optimizer with a learning rate of $5\times 10^{-3}$ to train our network. A first-in-first-out (FIFO) replay buffer stores the experiences of at most 5000 episodes, and a batch of 32 episodes are sampled from the buffer during the training phase. The target network is periodically updated every 200 episodes.

Our method and all the baselines involved in this paper are implemented based on the open-sourced codebase PyMARL \citep{samvelyan19smac}, which ensures the fairness of comparison. For the baselines (VDN \citep{sunehag2018value}, QMIX \citep{rashid2020monotonic}, DCG \citep{bohmer2020deep}, CASEC \citep{wang2021context}), we adopt the default hyperparameter settings provided by the authors.

The experiments are finished on NVIDIA RTX 2080TI GPU. The training time of SOP-CG and DCG are listed in Table \ref{table:training_time}. SOP-CG runs faster than DCG.

\begin{table}[h!]
    \centering
    \begin{tabular}{|c|c|c|c|}
      \hline
       &  \makecell[c]{Pursuit} & \makecell[c]{Tag} & \makecell[c]{1c3s5z} \\ \hline
      Environmental time cost & 2h & 6h & 1.5h\\ \hline
      DCG & 8.5h & 12.5h  & 7h \\ \hline
      SOP-CG ($\mathcal{G}_P$) & 6h & 9h & 5h \\ \hline
      SOP-CG ($\mathcal{G}_T$) & 8h & 9.5h & 6.5h \\ \hline
    \end{tabular}
    \caption{Time cost (hours) of training SOP-CG and DCG for 1M steps. Environmental time cost denotes the total time taken by the environment simulator and the training of the adversaries.}
    \label{table:training_time}
\end{table}

\section{Additional Results on MACO}
\subsection{Comparison to Fully Decomposed Value-Based Methods}

\begin{figure}[t]
    \centering
    \includegraphics[width=\linewidth]{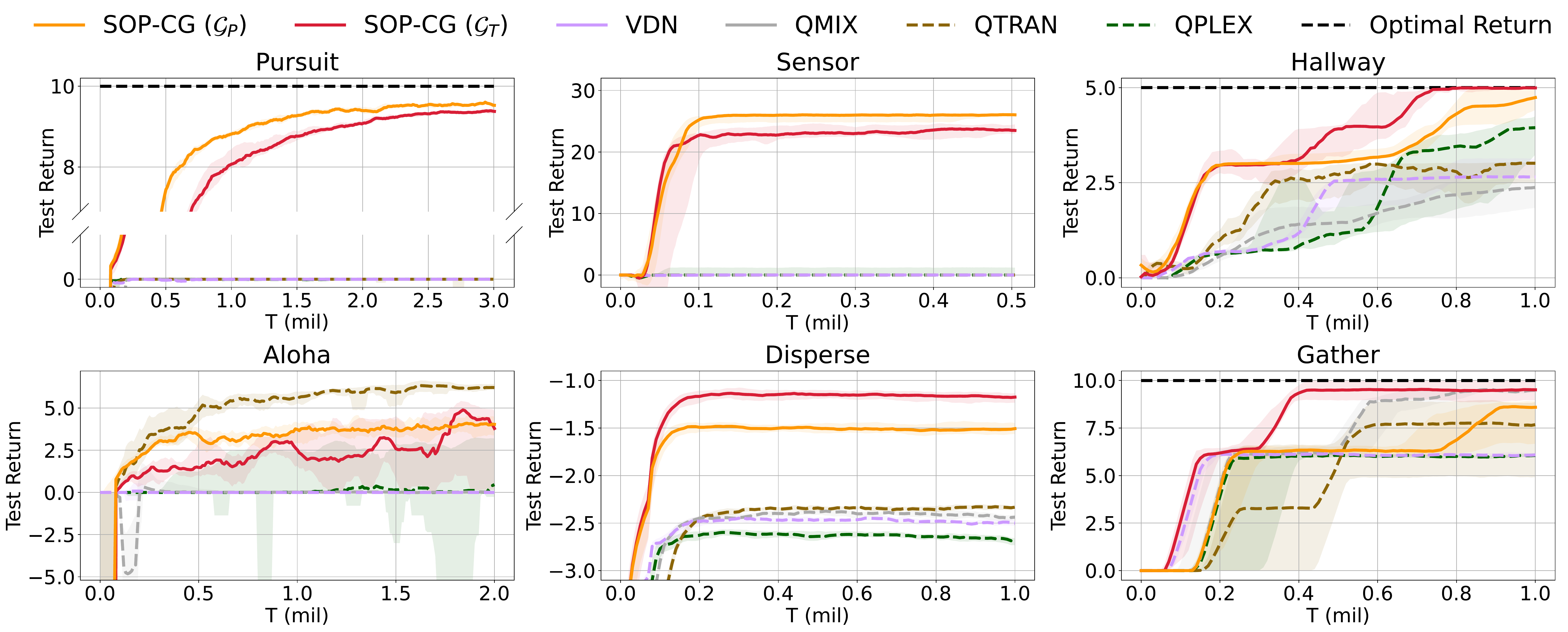}
    \vspace{-0.3in}
    \caption{Comparison to fully decomposed value-based methods on MACO.}
    \label{figure:maco_fullydec}
\end{figure}

We compare our algorithm with two additional fully decomposed value-based methods: QTRAN \citep{son2019qtran} and QPLEX \citep{wang2020QPLEX}. These methods achieve more expressive value function classes than VDN and QMIX. The performance is shown in Figure~\ref{figure:maco_fullydec}.

\subsection{Comparison to a Method Using the Underlying Problem Structure}
We test DCG with a fixed static graph that reflects the underlying problem structure of Aloha. As shown in Figure~\ref{figure:aloha}-Left, the peak performance of this method is similar to SOP-CG's. Notice that max-sum algorithm is not guaranteed to get accurate DCOP solutions on this underlying graph. If we further replace max-sum with an exhaustive search (which will take much higher time complexity), it achieves better peak performance than our method.

\subsection{Ablation of the Graph Selection Heuristic for $\mathcal{G}_T$}
\begin{figure}[t]
    \centering
    \begin{minipage}{0.4\linewidth}
        \centering
        \includegraphics[width=0.8\linewidth]{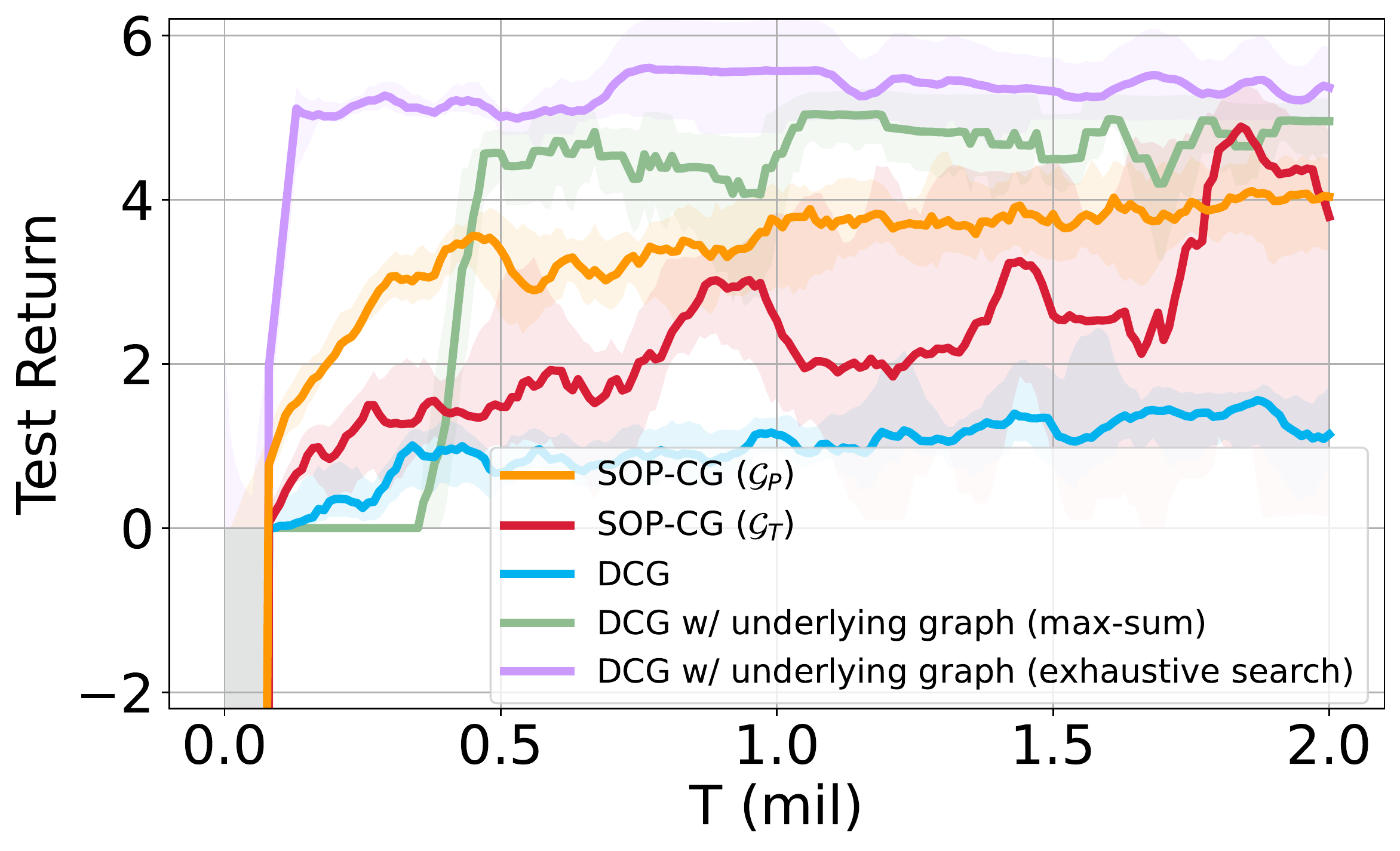}
    \end{minipage}
    \begin{minipage}{0.4\linewidth}
        \centering
        \includegraphics[width=0.8\linewidth]{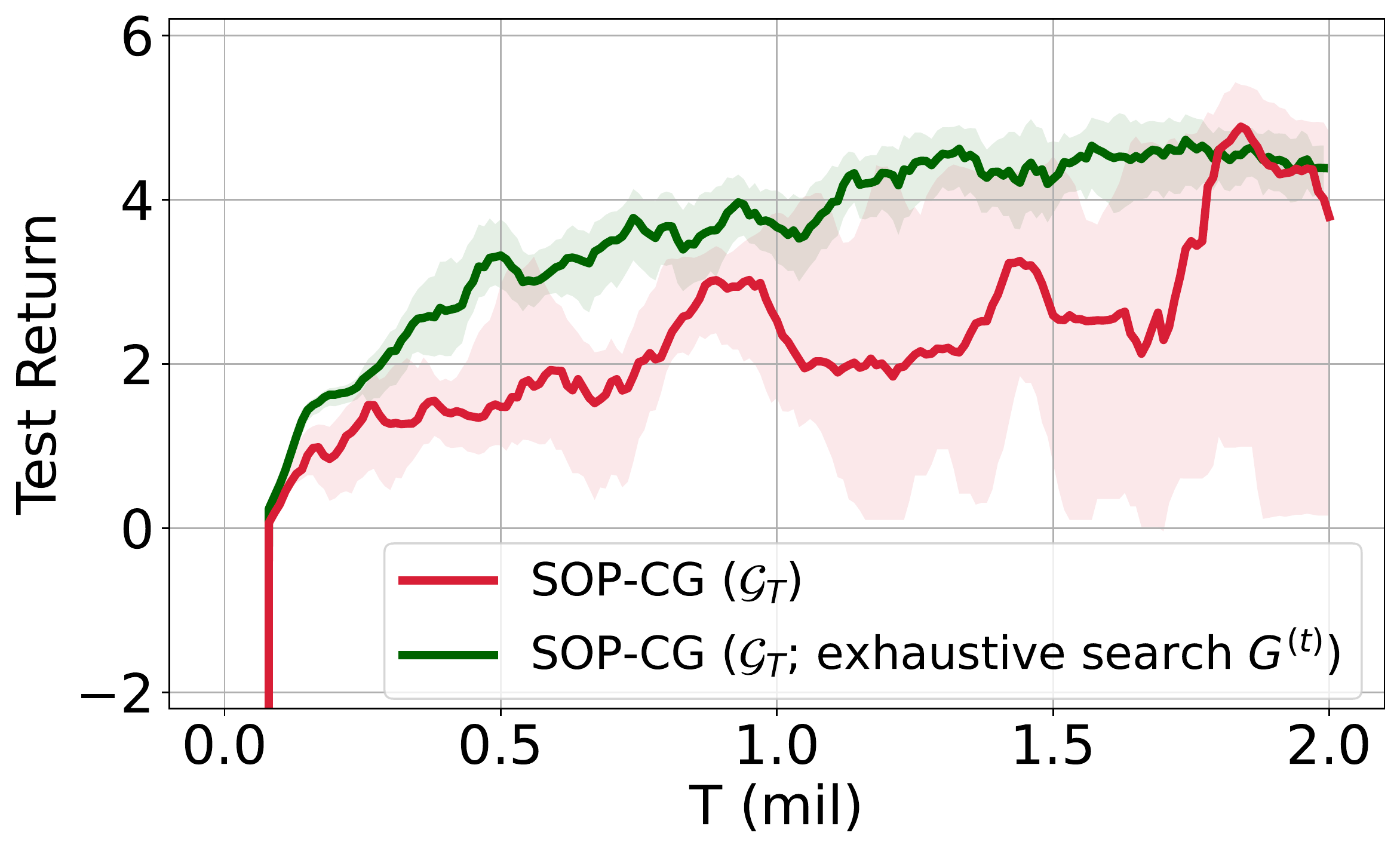}
    \end{minipage}
    \vspace{-0.1in}
    \caption{Additional results on Aloha. \textbf{Left}: Performance of DCG given a fixed static graph that reflects the underlying problem structure. \textbf{Right}: Performance of SOP-CG ($\mathcal{G}_T$) that selects $G^{(t)}$ by exhaustive search.}
    \label{figure:aloha}
\end{figure}

As mentioned in Appendix~\ref{appendix:algorithm}, we use a greedy heuristic to select a graph from class $\mathcal{G}_T$. To understand the performance of this heuristic, we replace it with an exhaustive search (which will take much higher time complexity) and test it on Aloha. As shown in Figure~\ref{figure:aloha}-Right, the peak performance of this implementation is similar to the one with the greedy heuristic, but the precise graph selection brings better stability.

\section{Connection to a Recent Work on Hyper-Graphs}
We note that a recent work on the expressiveness of hyper-graphs \citep{yoon2020much} is relevant to our work. The connections (i.e., the differences and relevance) between \citet{yoon2020much} and our paper are summarized as follows:
\begin{itemize}
    \item \textbf{Different metric of expressiveness.} \citet{yoon2020much} investigates the functionality of graph expressiveness in terms of the order of hyper-edges. By contrast, our paper focuses on the expressiveness of different graph topologies.
    \item \textbf{Similar motivation.} Both Yoon et al. and our paper aim to study the trade-off between the graph expressiveness and the complexity of solving downstream tasks. In our paper, the downstream task specifically refers to the DCOPs induced by graph-based value factorization.
    \item \textbf{Different workflow and contributions.} \citet{yoon2020much} establish a systematic experimental study to provide an empirical guideline on how to trade-off between the graph expressiveness and downstream costs. In comparison, our work proposes an algorithm to break this trade-off for improving MARL performance. Our algorithm is able to improve the function expressiveness without largely increasing the time complexity of downstream DCOPs.
\end{itemize}

Despite the expressiveness metric being different, several conclusions reached by \citet{yoon2020much} can be observed in our experiments. More specifically, the key findings of \citet{yoon2020much} (namely, three points, i.e, \textit{Diminishing returns}, \textit{Troubleshooter}, and \textit{Irreducibility}) correspond to the following observations in our paper:

\begin{itemize}
 	\item \textbf{Diminishing returns.} When the task coordination structure is simple, a concise graph topology (e.g., pairwise matching) can achieve comparable performance with topologies using larger edge sets. As presented in Fig.~\ref{figure:maco}, SOP-CG ($\mathcal{G}_P$) performs better than SOP-CG ($\mathcal{G}_T$) and DCG in Pursuit, since pairwise matching is sufficient to express the underlying coordination relations in this task. (Recall that $\mathcal{G}_P$ uses pairwise matching and $\mathcal{G}_T$ uses tree-based topology.)
 	\item \textbf{Troubleshooter and Irreducibility.} When the task requires a complicated coordination structure, the graph topology needs to have higher function expressiveness. This finding can also be seen in Fig.~\ref{figure:tag}. SOP-CG ($\mathcal{G}_T$) significantly outperforms SOP-CG ($\mathcal{G}_P$) in Tag where pairwise matching cannot express the underlying coordination relations.
\end{itemize}

\end{document}